\documentclass{article}
\usepackage[utf8]{inputenc}
\usepackage[a4paper]{geometry}
\usepackage{natbib}
\usepackage{graphicx}
\usepackage{subcaption}
\usepackage{setspace}
\usepackage[hidelinks]{hyperref}

\title{Semiotically-grounded distant viewing of diagrams:\\insights from two multimodal corpora}
\author{Tuomo Hiippala \& John A. Bateman}
\date{}

\begin{document}

\maketitle

\section{Introduction}

Whether taking place via an external medium or in face-to-face interaction, communication is naturally multimodal: that is, making and exchanging meanings involves combining multiple modes of expression in a coordinated, goal-oriented manner. There is currently growing interest in multimodal communication across various fields of research, including the digital humanities. It is then natural to consider more closely whether contemporary theories of multimodality can support the kinds of large-scale analyses commonly pursued in digital humanities and if so, to what extent. Although the field of multimodality is increasingly oriented towards empirical analysis, compiling multimodal corpora to support such analyses is still highly labour-intensive. More extensive use of computational techniques is thus a clear priority. 

In this article we consider the potential benefits of combining contemporary accounts of multimodality, computational methods, and research orientations from digital humanities with respect to one extremely common mode of expression, namely diagrams. We argue that a dialogue between multimodality research and digital humanities may then prove mutually beneficial. We approach this by drawing on two recent diagram corpora developed for different purposes. While the first corpus was developed to support research on artificial intelligence for tasks such as automatic diagram understanding, the second corpus builds multiple layers of annotation on top of the first corpus to create a resource for studying how diagrams communicate multimodally. Against this backdrop, we ask: how can modern multimodality theory inform the design and analysis of multimodal datasets and corpora? To answer this question, we critically evaluate the two datasets from the perspective of multimodality theory and the digital humanities notion of `distant viewing', supporting the argument using quantitative and computational methods.

\section{Multimodality research and the digital humanities}
\label{sec:mmdh}

Multimodality research is an emerging discipline that examines how communication builds on appropriate combinations of `modes' of expression, such as natural language, illustrations, drawings, photography, gestures, layout and many more \citep{wildfeueretal2020}. Although now widely acknowledged as an inherent feature of human communication, multimodality is not always understood in the same way across the diverse fields of study where the concept has been picked up. These fields include, among others, text linguistics, spoken language and gesture research, conversation analysis \citep{mondada2019}, human-computer interaction \citep{oviattcohen2015} and, last but not least, digital humanities \citep{svensson2010}. To consolidate these perspectives, \citet{batemanetal2017} propose a general framework for multimodality that extends beyond previous approaches by offering a common set of concepts and an explicit methodology for supporting empirical research regardless of the `modes' and materialities involved.

Modern multimodality theory has developed a battery of interrelated theoretical constructs to support descriptive and empirical analyses of complex communicative situations and artefacts \citep{stoeckl2020}. Several core concepts, including in particular semiotic mode \citep{bateman2011,kress2014}, medium \citep{batemanetal2017,bateman2017} and genre \citep{bateman2008,hiippala2015a}, theorise in detail how individual forms of expression are structured and what enables them to effectively combine and co-operate with each other across a wide range of communicative contexts and situations. Despite this comprehensive theoretical apparatus, which is now mature enough to be brought into productive discussion with established fields of study such as media archeology \citep{thomas2020a}, literacy \citep{Jewitt2008}, ethnography \citep{kress2011}, and others, the lack of large annotated corpora stands in the way of refining multimodality theory through empirical research \citep{thomas2020b}. 

Within linguistics, the provision of ever larger collections of authentic language use established corpus linguistics as a major pillar of research. The corresponding treatment of multimodal data, and particularly \emph{static} multimodal data,  still lags very much behind, constituting a major bottleneck in evaluating and refining the theoretical constructs proposed. Whereas it is common for a range of automatic processing techniques to be applied to linguistic corpora, the possibilities for multimodal data remain limited. As a consequence, multimodal annotation frameworks are slow to develop and require expert annotators. Most current multimodal corpora consequently remain small and thus resemble curated collections rather than true corpora in the linguistic sense of the term. Although this problem applies in principle both to face-to-face interaction \citep{huang2020} and multimodal documents \citep{waller2017}, substantial progress is now being made for some forms of audiovisual multimodal data \citep[e.g.,][]{Steen-etal2018-datadriven}. The situation for multimodal documents is very different and it is this area that we focus on here.

Parallel to the aforementioned developments in multimodality research, there is a growing interest in the large-scale analysis of `visual' communication within the field of digital humanities as well \citep[see e.g.][]{langommer2018, Heftberger2018, arnoldtilton2019, weverssmits2019, munsterterras2020, smitsros2020}. \citet{arnoldtilton2019} propose a framework for \emph{distant viewing}, arguing that such approaches are needed to counterbalance the strong textual orientation in digital humanities, which excludes a wealth of non-linguistic phenomena that are traditionally of interest to the humanities. Drawing on foundational work in semiotics in the tradition of Saussure and Barthes, \citet[i4]{arnoldtilton2019} observe that:
\begin{quote}
``in order to view images computationally, a representation of elements contained within the visual material -- a code system in semiotics or, similarly, a metadata schema in informatics -- must be constructed.''
\end{quote}
Put differently, \citet{arnoldtilton2019} emphasise the need to impose analytical control over visual material, just as ``the explicit code system of written language'' allows imposing structure on textual corpora by establishing units of analysis (e.g. tokens or parts of speech) and their interrelations (e.g. syntax) \citep[i5]{arnoldtilton2019}. However, just which analytical units should be defined for modes of communication other than language and at what level of granularity remains an open and hotly contested question.

Providing such a capability by building on more traditional semiotic notions of `code' is, however, unlikely to succeed for visual materials. As argued more extensively in \citet[21--22]{bateman2017b}, any model that treats communication in terms of  an exchange of meanings in a process of encoding/decoding according to some fixed, static code can readily be shown to be inadequate. Early debates on this `fixed code fallacy' focused mainly on language and excluded other modes of communication \citep[232]{cobley2013}, which may explain why early research on multimodality already turned towards developing the concept of `semiotic mode' as an alternative.  Semiotic modes, of which language is taken as just one alongside many others, are assumed to emerge and  be shaped through social interaction within a community of users \citep{kressvanleeuwen1996, kressvanleeuwen2001}. However, as \citet[22]{bateman2017b} points out, early definitions of mode were not robust enough to support empirical research, leading common conceptions of mode and code to gradually became indistinguishable as semiotics itself advanced beyond the encoding/decoding model of communication \citep[cf.][231]{cobley2013}.

\begin{figure*}[h!]
\centering
\begin{subfigure}[t]{0.48\textwidth}
\includegraphics[width=1\linewidth]{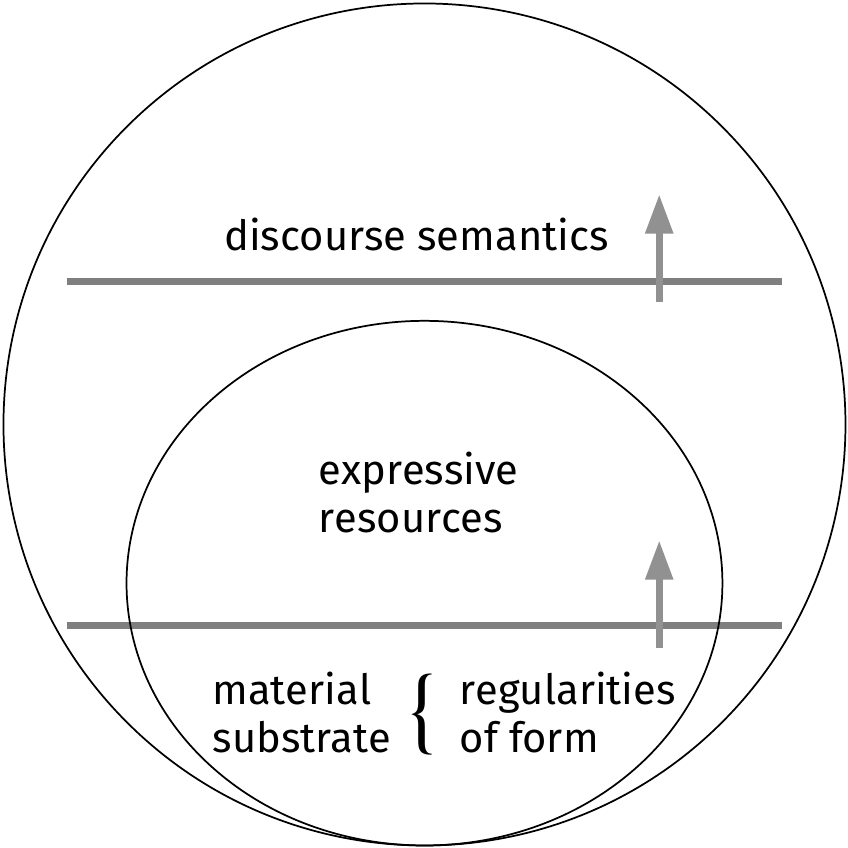}
\caption{A theoretical model of a semiotic mode}
\label{fig:semiotic_mode}
\end{subfigure}
~ 
\begin{subfigure}[t]{0.48\textwidth}
\includegraphics[width=1\linewidth]{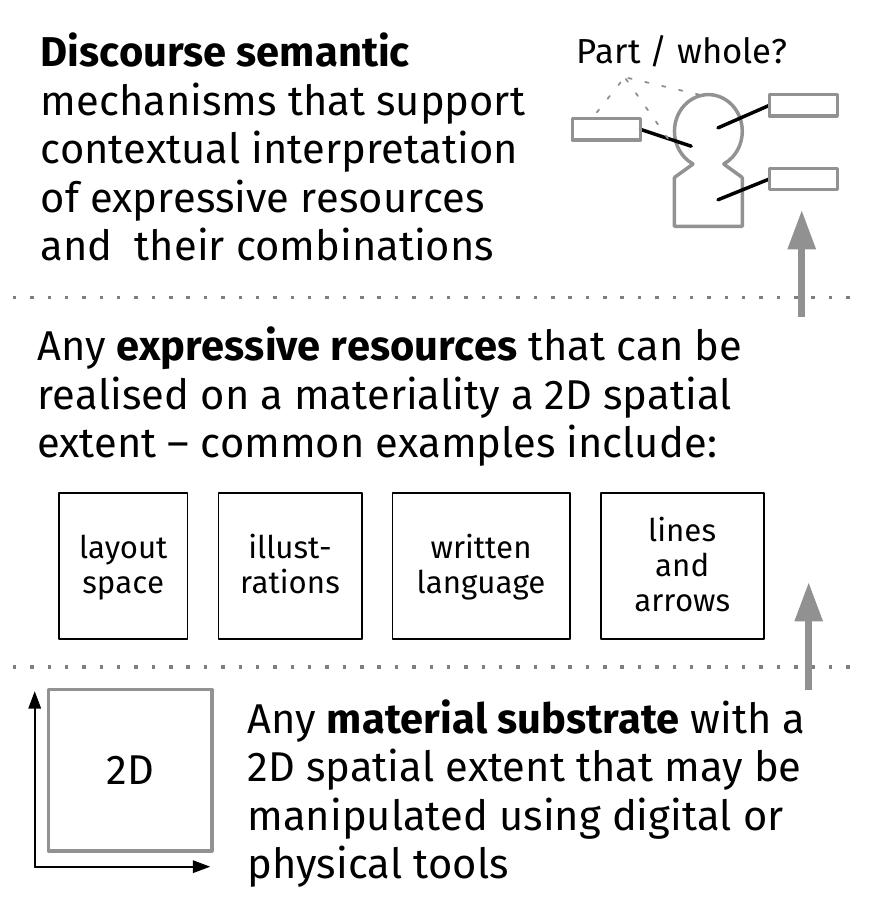}
\caption{A characterisation of the diagrammatic mode}
\label{fig:diag_mode}
\end{subfigure}
\caption{The concept of a semiotic mode and its application to diagrams}
\label{fig:sem_modes}
\end{figure*}

\citet{bateman2011} addresses previous shortcomings with the definition of semiotic mode by proposing a formal account consisting of three semiotic strata, visualised in Figure \ref{fig:semiotic_mode}. Each stratum is a prerequisite for a fully developed semiotic mode. Starting from the bottom of the inner circle, all semiotic modes must work with respect to some \emph{materiality} that can be manipulated intentionally for communicative purposes. Such traces of manipulation must reflect formal distinctions that are pertinent for \emph{expressive resources} available within the semiotic mode, as exemplified by differences in form between written language and line drawings, which allow us to distinguish between these resources. The expressive resources are assumed to be subject to a paradigmatic organisation that allows making selections among them and combining them into larger syntagmatic organisations. Third, the expressive resources and their combinations are mobilised in the service of communication by a corresponding \emph{discourse semantics}, which supports the contextual interpretation of selections made within expressive resources, and whose operation we illustrate in a moment. This general model places no restrictions on the kinds of materiality that may be employed; for current purposes, however, we focus on static materialities with a 2D spatial extent, such as a sheet of paper or a static display presented on a screen.

Figure \ref{fig:diag_mode} exemplifies the application of this general model to what may be tentatively called the `diagrammatic mode' -- a semiotic mode underlying all kinds of diagrammatic representations \citep{hiippalabateman2020}. Beginning from the bottom, diagrams always require a materiality with a 2D spatial extent -- any materiality capable of doing the job will do, which is why one encounters diagrams everywhere from public signs to newspapers and school textbooks. Consequently, the diagrammatic mode can theoretically draw on all expressive resources that can operate with a 2D spatial materiality, although which expressive resources are actually mobilised and the choices made within them are largely motivated by genre, that is, the communicative goals set for the diagram in its context of use \citep{KostelnickHassett2003,lemke2005}. 

It is an intrinsic property of this model that multiple expressive resources such as written language, illustrations and photographs may \emph{naturally} co-occur with each other in diagrams, thus avoiding committing to arbitrary divisions between `verbal' and `visual' or `text' and `image' \citep{bateman2014b}. This perspective is obviously carried over to mass media, which regularly deploy multiple semiotic modes \citep[124]{batemanetal2017}. Finally, discourse semantics guides the interpretation of expressive resources and their combinations in context. For diagrams, resolving the resulting discourse relations relies on formal cues such as spatial placement of elements or connections realised using lines and arrows in combination with world knowledge \citep{watanabenagao1998, alikhanistone2018}.

This brief description illustrates the extent to which modern multimodality theory can explicate how semiotic modes operate more generally: describing the characteristics of each stratum of a given semiotic mode is then an issue that warrants empirical research. Conversely, this also shows just how much of this complexity is missed when operating with pre-theoretical distinctions such as `text' and `image'. Although computational analyses of page-based media are already advancing beyond such dichotomies, as \citet{weverssmits2019} have shown by training convolutional neural networks to distinguish between instances of illustrations, photographs and other semiotic modes in historical newspapers, we argue that advancing this effort within digital humanities would benefit still further from the input of multimodality theory. 

As a form of `applied semiotics' that seeks a close relationship between theory and data \citep{batemanhiippala2021}, multimodality theory is well-positioned to provide a foundation for characterising the diverse range of communicative artefacts and situations studied within digital humanities \citep{bateman2017b}. Theories of multimodality have already been used to guide the application of computational methods to both filmic \citep{batemanetal2016} and page-based media \citep{ohalloranetal2017b}, but much remains to be done in terms of applying computational methods in a way that respects the complexity of multimodal communication. While we are not suggesting that all studies that use computational methods should perform full-blown multimodal analyses for each semiotic mode encountered, we do  encourage approaches where multimodality theory is used to determine the analytical granularity needed. This allows the selection of targets of descriptions and their respective granularity to be derived systematically and on the basis of a developing body of theory. There are then both theoretical and practical reasons for adopting this approach when constructing multimodal corpora. 

As noted by \citeauthor{arnoldtilton2019} above, appropriate metadata schemes are needed for organising data when constructing large collections or corpora for computational analyses. In computer science, however, the definition of `modality' (the term preferred over `mode') is strongly aligned with the senses: our ability to see, hear, touch and use natural language. A clear distinction between these sensory modalities is often taken for granted in research on computer vision, audio signal processing and natural language processing, and these are then the sources for corresponding metadata schemes. The corresponding fields have been largely confined to their own `problem spaces', even though these are now converging towards multimodality in tasks such as machine translation \citep{sulubacaketal2020}. Nevertheless, work broadly continues applying definitions based on sensory modalities. Restricting approaches within sensory channels is now receiving considerable critique in humanities-oriented multimodality theory for several reasons. In particular, predefining modalities ahead of analysis simply on the basis of perceptual properties makes it difficult to identify the actual semiotic contributions to meaning construction being made. As argued extensively in \citet{batemanetal2017}, such contributions regularly extend across sensory channels and need to be teased out empirically: one cannot assume their individual characteristics in advance \citep[17--18]{bateman2011}. It is crucial for a channel to be opened up between the semiotic distinctions being made and material distinctions, rather than assuming that sensory perception will provide necessary and appropriate segmentations. 

These challenges and limitations may be exemplified in light of the diagrammatic mode shown in Figure \ref{fig:diag_mode}. First, diagrams are clearly not aligned with a single traditional modality, as they cross-cut both `vision' and `language'. Second, what makes diagrams different from other combinations of a similar nature, such as photographs with embedded or overlaid text, remains an open question. Assumptions of similarity concerning expressive resources \emph{within} a single sensory modality are common in computer vision research, in which objects of analysis are often reduced to mere carriers of content. This is insufficient, however. \citet[9]{haehnetal2019}, for example, report that models trained on photographs do not generalise well to diagrammatic representations without further training even though they are both clearly `visual'. They consider this finding surprising given prior comparisons between artificial neural networks and the human visual cortex, which assume that \emph{visual perception} suffices for reasoning about both photographs and diagrams. 

From the perspective of multimodality theory, however, the differences between photographs and diagrams are rather evident: diagrams differ radically in terms of their expressive resources and discourse semantics. Diagrams are compositional, that is, they can be broken down into component parts, which may be realised using multiple expressive resources and combined into discourse structures that work towards a shared communicative goal. This allows diagrams to represent abstract concepts and phenomena that are not limited to specific slices of time and space, which stands in strong contrast to the nature of photographs \citep[cf. e.g.][]{alikhanistone2018, greenberg2018}. This demonstrates how it is always essential to consider material distinctions in terms of the particular semiotic modes that they are operating with respect to. It is then precisely these semiotic modes that deliver appropriate metadata schemes for characterising corresponding objects of analysis.  

\section{Insights from multimodal diagram corpora}

Having introduced the concept of a semiotic mode and how appropriate metadata schemes may be derived for individual semiotic modes through empirical research, we now turn to examine two recent diagram corpora from this perspective. These corpora originate in two different fields of research, namely artificial intelligence and multimodality research, whose disciplinary foci are reflected in the metadata schemes used to describe the diagrams and their structure. 

\subsection{AI2D -- a dataset for computational processing of diagrams}
\label{sec:ai2d}

The first dataset is the \emph{Allen Institute for Artificial Intelligence Diagrams} dataset (AI2D), which was developed to support research on visual question answering, automatic diagram understanding and other computational tasks involving diagrams in the field of artificial intelligence \citep{kembhavietal2016}. The AI2D dataset contains 4903 diagrams that represent 17 topics in elementary school natural sciences, ranging from life and carbon cycles to human physiology and food webs. The dataset models four types of diagram elements: text, arrows, arrowheads and blobs. Whereas the first three categories are rather self-explanatory, `blobs' is a technical term that refers to all visual expressive resources deployed in AI2D diagrams, such as line drawings, illustrations and photographs \citep{tverskyetal2000}. To summarise, AI2D relies on pre-defined categories for diagram elements, which are assumed to be known in advance.

Placing objects into pre-defined categories is a common strategy in crowdsourcing annotations for computer vision research \citep{kovashkaetal2016}, but faces problems similar to those encountered when defining analytical categories ahead of actual analysis \citep[18]{bateman2011}. To reiterate, if the expressive resources of a semiotic mode are assumed to be known in advance, then it becomes difficult to explicate just what a given semiotic mode does with its material substrate, that is, what kinds of `regularities of form' or expressive resources are made available by the semiotic mode and what can be done with them in terms of communication. As we will see below, decisions related to modelling expressive resources are also propagated to the stratum of discourse semantics, which complicates its description as well (cf. Figure \ref{fig:sem_modes}).

Each diagram in the AI2D dataset is provided with several types of description. All instances of text, arrows, arrowheads and blobs were first segmented from the original diagram layout by crowdsourced workers on Amazon Mechanical Turk\footnote{https://www.mturk.com}. The diagram elements identified during \emph{layout segmentation} provide a foundation for a \emph{Diagram Parse Graph} (DPG), which represents the diagram elements as nodes, whereas edges define the semantic relations that hold between the elements. These semantic relations are described using ten relation definitions drawn from the framework proposed for diagrammatic representations by \citet{engelhardt2002}. The following examples illustrate the application of the AI2D annotation schema using a single diagram from the dataset. 
\begin{figure}[h!]
\centering
\includegraphics[width=0.65\textwidth]{4210.png}
\caption{Diagram \#4210 in the AI2D dataset}
\label{fig:orig_diag}
\end{figure}
The original diagram in Figure \ref{fig:orig_diag} represents a rock cycle, that is, transitions between different types of rock, using a combination of an illustration (a cross-section) whose parts are described using written language. These parts set up the stages of the rock cycle, which are then related to one another using arrows and written language.

\begin{figure}[h!]
\centering
\includegraphics[width=0.65\textwidth]{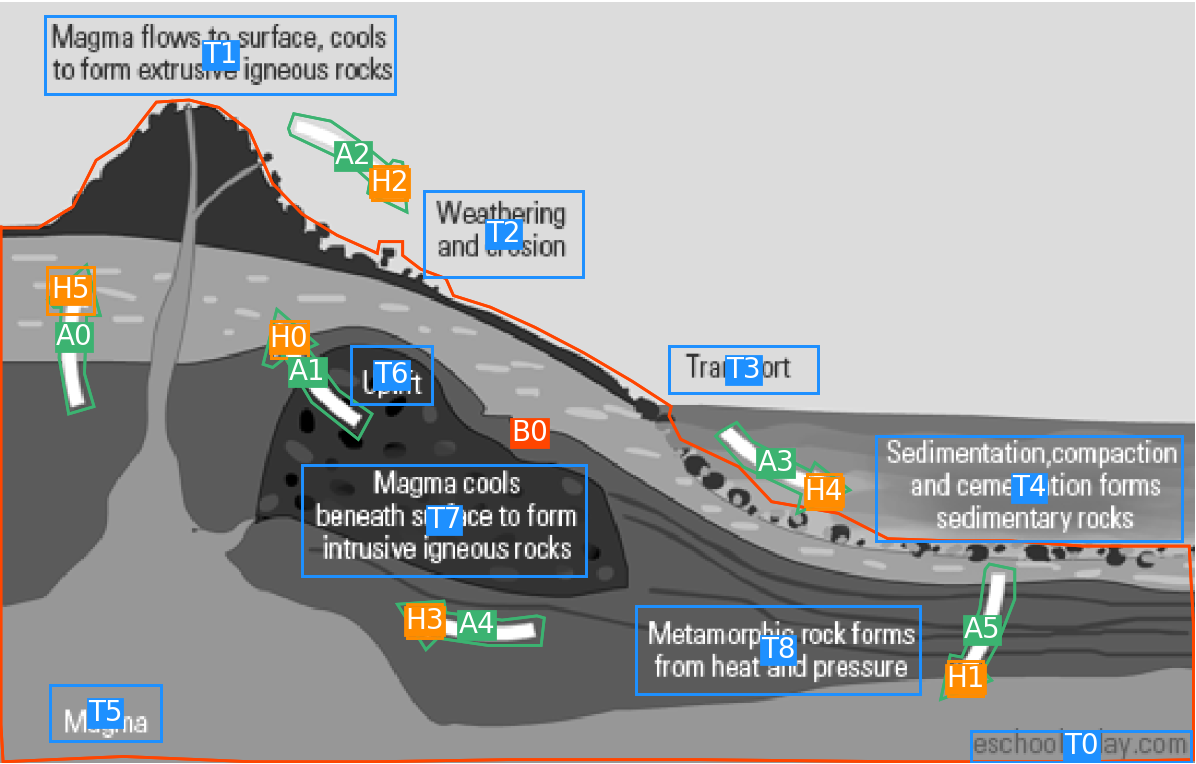}
\caption{Layout segmentation. The diagram image has been converted into grayscale to highlight the crowdsourced layout segmentation. Each layout segment is coloured according to diagram element type (blue: text; red: blob; arrow: green; arrowhead: orange) and assigned a unique identifier, which are laid out on top of the bounding boxes and polygons.}
\label{fig:lay_seg}
\end{figure}

As pointed out above, the crowdsourced workers were first requested to identify instances of diagram elements during layout segmentation. Figure \ref{fig:lay_seg} shows that text blocks and arrowheads were segmented using rectangular bounding boxes, whereas more complex shapes for arrows and various types of graphics were segmented using polygons. The layout segmentation illustrates well how crowdsourced annotators tend to segment diagrams to quite uneven degrees of detail.
The entire cross-section of a volcano in Figure \ref{fig:lay_seg} is assigned to a single blob (B0), although arguably a more accurate description would be to segment separate parts of the cross-section, such as magma and various layers of rock. Demarcating such meaningful regions on material substrates with a 2D spatial extent is a hallmark feature of illustrations and other graphic expressive resources, which diagrams regularly deploy for communicative purposes in combination with written text \citep{richards2017}. As we will show shortly below, ignoring this feature places serious limitations on the possible description of discourse semantics in diagrams. 

\begin{figure}[h!]
\centering
\includegraphics[width=0.65\textwidth]{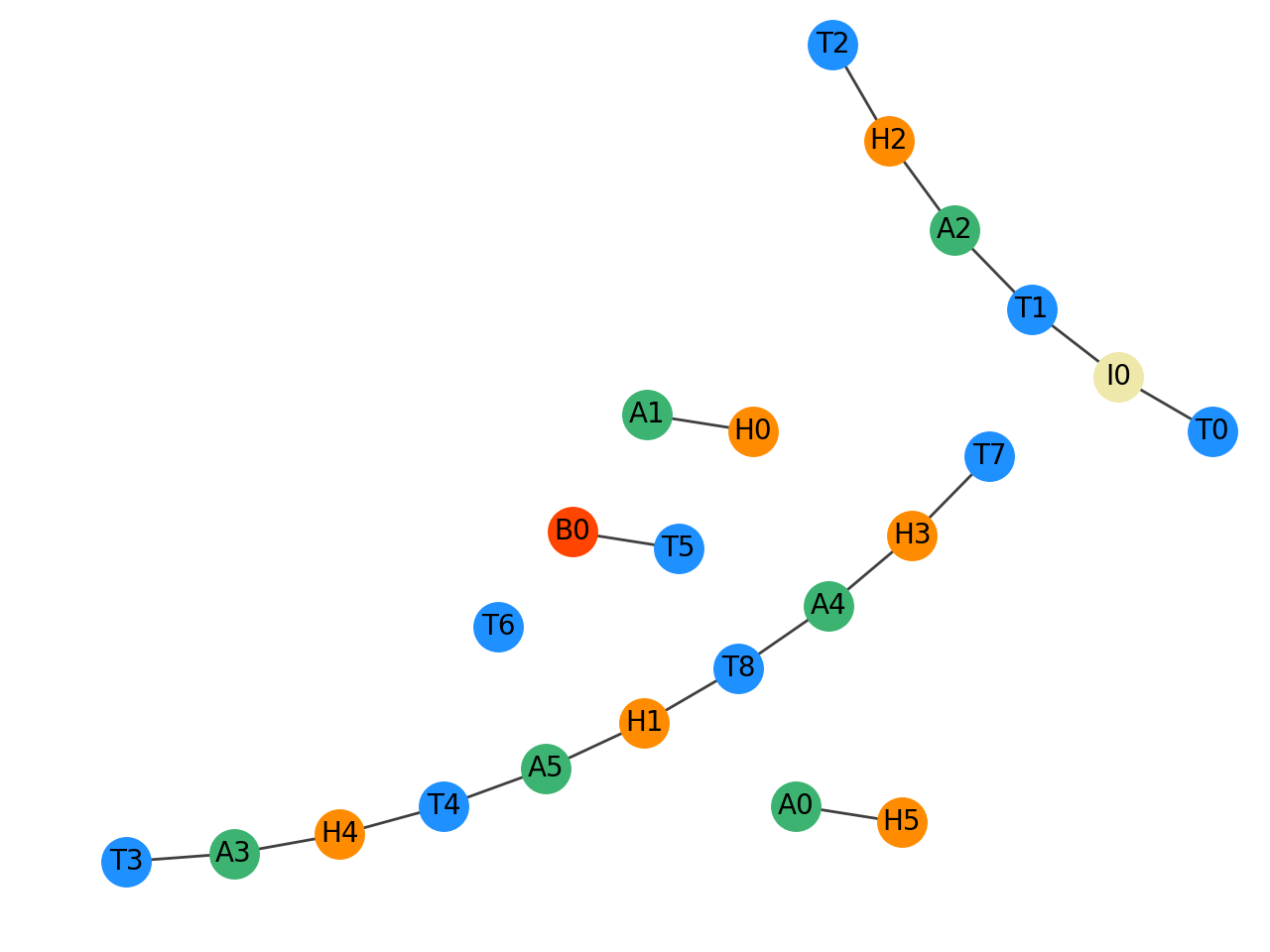}
\caption{Diagram Parse Graph. The identifiers are carried over from the layout segmentation.}
\label{fig:dpg}
\end{figure}
The edges of the DPG shown in Figure \ref{fig:dpg} carry relations such as \textsc{arrowHeadTail} between arrow A2 and arrowhead H2 in the upper part of Figure \ref{fig:lay_seg}. The arrow A2, in turn, acts as a connector in an \textsc{interObjectLinkage} relation between text blocks T1 (`Magma flows to surface ...') and T2 (`Weathering and erosion'). As these relations exemplify, the relations drawn from \citet{engelhardt2002} cover \emph{local} relations that hold between diagram elements positioned close to each other or connected using arrows or lines \citep[239]{kembhavietal2016}, but they neglect the relations needed to describe the \emph{global} organisation of the diagram, that is, relations between units that are made up of multiple elements. This is not a shortcoming of Engelhardt's \citeyearpar{engelhardt2002} framework, but rather its extremely limited application in the AI2D dataset.

Crowdsourcing graph-based descriptions of diagrams without a descriptive schema that covers both local and global discourse structures is undoubtedly a challenging task, which may explain why isolated nodes and multiple connected components are commonly found in AI2D DPGs (see, e.g., the isolate T6 and five connected components in Figure \ref{fig:dpg}). Although the diagram in Figure \ref{fig:dpg} shows a rock cycle, the cyclic nature of this phenomenon is not reflected by the structure of the DPG at all, even though the AI2D annotation schema does in principle provide the relation definitions necessary for describing such cycles, including \textsc{interObjectLinkage} and \textsc{intraObjectRegionLabel} \citep[239]{kembhavietal2016}.

Here the shortcoming is caused by insufficient detail in the layout segmentation. The crowdsourced annotators were not instructed to decompose instances of graphic expressive resources into meaningful regions, although such resources are commonly deployed in diagrams precisely due to their capability to demarcate  regions `in the world' or described phenomena. The blob B0, which covers the entire cross-section, is consequently not segmented into its component parts, which are indicated by labels such as `Magma' (T5) and `Metamorphic rock forms from heat and pressure' (T8) which pick out particular regions of the cross-section through visual \emph{containment} \citep[47]{engelhardt2002} to set up the stages of the rock cycle. Figure \ref{fig:dpg} shows that only the label T5 (`Magma') is related to blob B0 through a relation of \textsc{intraObjectRegionLabel} (note that the edges that stand for particular relations are not labelled in Figure \ref{fig:dpg}). Because the cross-section (B0) constitutes a single unit, otherwise applicable relations such as \textsc{intraObjectRegionLabel} cannot be used to map the labels (e.g. T4, T6, T7, T8) to the corresponding regions of the cross-section. Consequently, the regions that would be needed to represent the cyclic structure are not available in the inventory of annotated diagram elements.

These challenges in decomposing diagrammatic representations relate to the well-known problem of identifying analytical `units' in any visually-driven media. \citet{batemanwildfeuer2014b} consider this issue for comics and argue for a discourse-based approach to identifying analytical units, whereby the discourse organisation of some larger unit (e.g. a panel in a comic or an entire diagram) help determine which elements are to be picked up for interpretation in a given context. In other words, the stratum of discourse semantics simultaneously supports \emph{decomposing} larger units into their component parts and \emph{resolving} their potential interrelations. In contrast to often criticised attempts to impose fixed units or parts on visual materials, in the discourse-based account segmentation is always pursued with the goal of maximising \emph{discourse coherence} via abductive reasoning \citep[377]{batemanwildfeuer2014b}, that is, updating discourse interpretations as additional evidence becomes available, while striving to form the most plausible explanation for a particular constellation of expressive resources. For this reason, it will often be more effective not to operate with a pre-defined inventory of elements (i.e., defining units bottom-up), but instead to allow the inventory of relevant elements to change dynamically as interpretations are made and updated (top-down). Such mechanisms are inherent to the stratum of discourse semantics \citep{Bateman2020-discsem}. 

\subsection{AI2D-RST -- adding discourse semantics to AI2D diagrams}
\label{sec:ai2d-rst}

The second corpus of diagrams, AI2D-RST, features multiple layers of expert annotations built on top of the crowd-sourced layout segmentations from the AI2D dataset \citep{hiippalaetal2019-ai2d}. The AI2D-RST corpus covers 1000 diagrams from AI2D. AI2D-RST attempts to provide a more comprehensive description of discourse structure in diagrams, whose metadata scheme is directly derived from the assumed discourse semantics of the diagrammatic mode. For this purpose, AI2D-RST employs a formalism that is frequently used to describe the operation of discourse semantics, that of Rhetorical Structure Theory (RST), which was originally developed as a theory of text organisation and coherence in the 1980s \citep{mannthompson1988}. RST attempts to describe why well-formed texts appear coherent, or why individual parts of a text appear to contribute towards a common communicative goal. The RST framework was extended for the description of multimodal documents in the early 1990s \citep{andrerist1995} and has been used as a model of discourse semantics for various modes and media, ranging from bird field guides and tourist brochures to scientific publications and product packaging \citep{bateman2008,taboadahabel2013,thomas2014,hiippala2015a}. 

\begin{figure*}[h!]
\centering
\begin{subfigure}[t]{0.48\textwidth}
\includegraphics[width=1\linewidth]{segmentation_4210.png}
\caption{Layout segmentation from AI2D}
\label{fig:lay_seg2}
\end{subfigure}
~ 
\begin{subfigure}[t]{0.48\textwidth}
\includegraphics[width=1\linewidth]{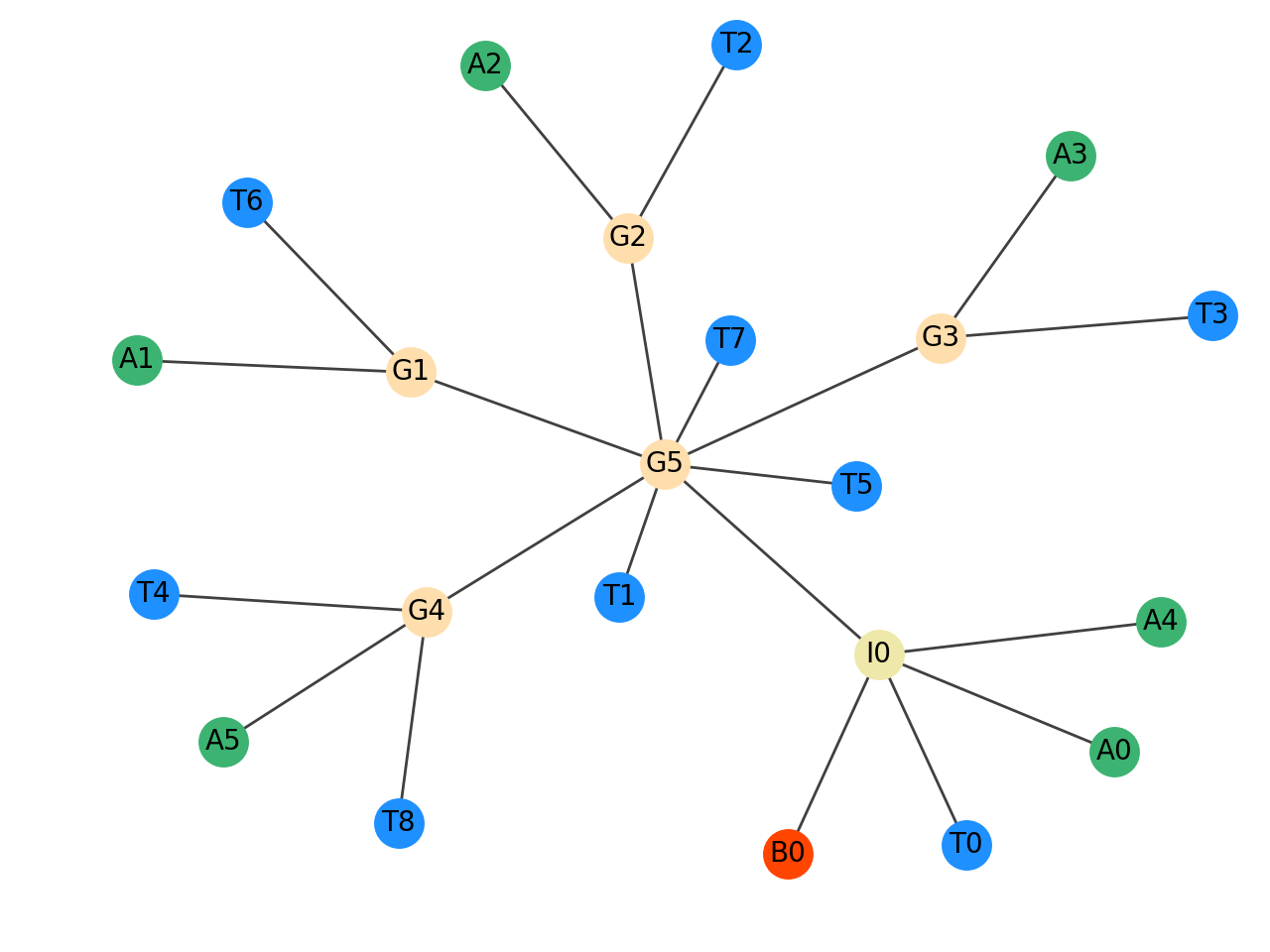}
\caption{Grouping graph}
\label{fig:gg}
\end{subfigure} \\
\begin{subfigure}[t]{0.48\textwidth}
\includegraphics[width=1\linewidth]{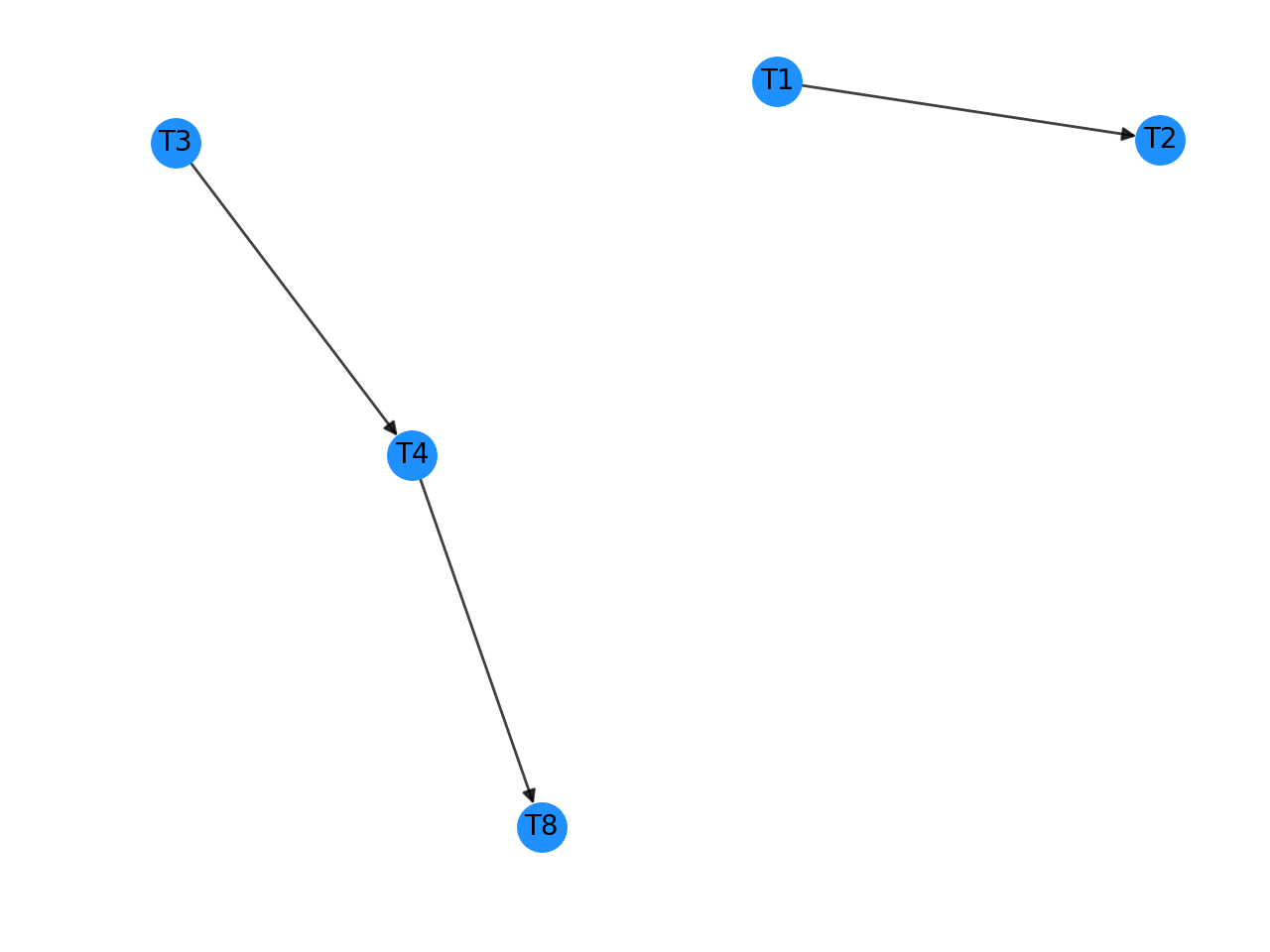}
\caption{Connectivity graph}
\label{fig:cg}
\end{subfigure}
~ 
\begin{subfigure}[t]{0.48\textwidth}
\includegraphics[width=1\linewidth]{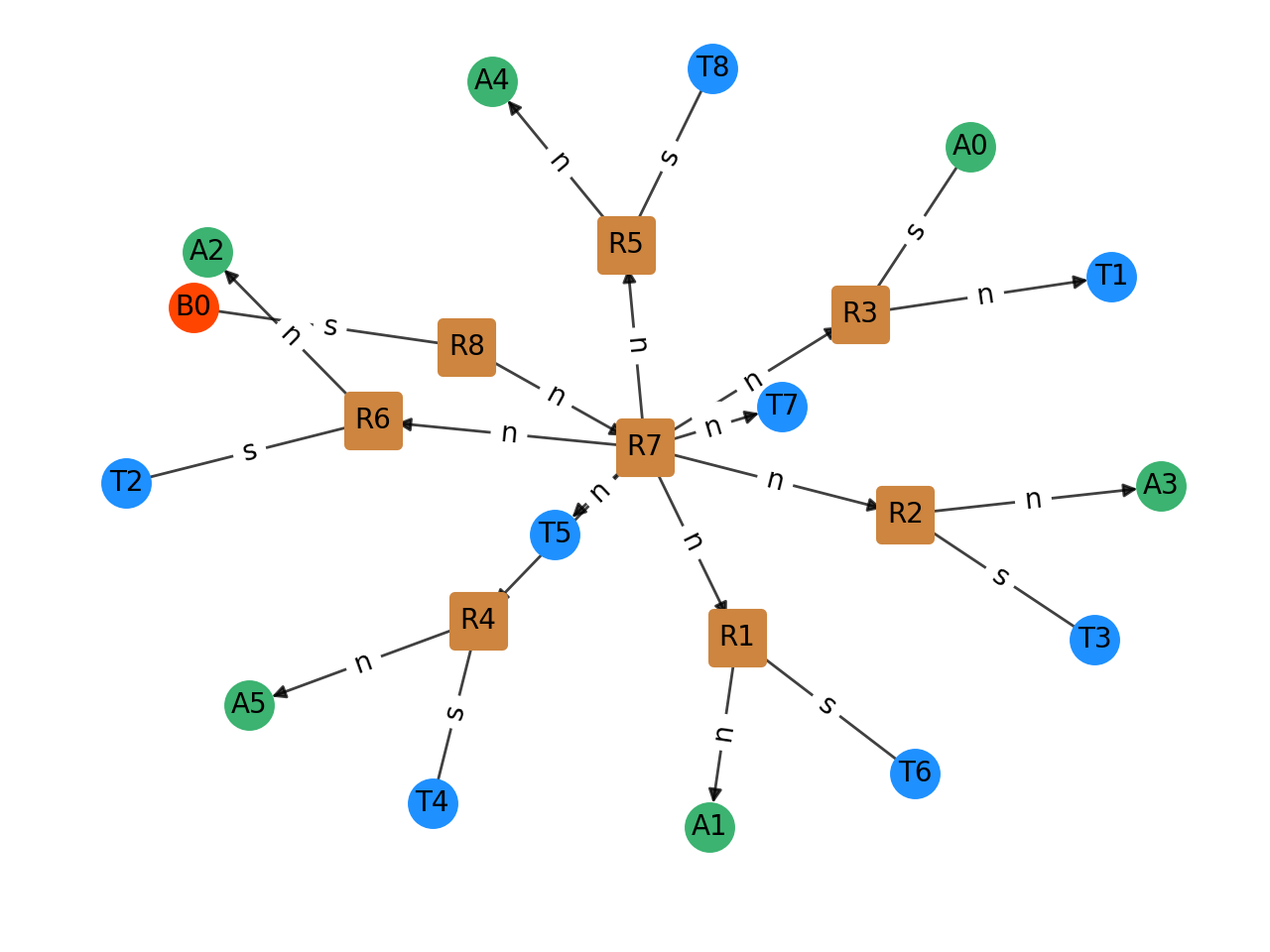}
\caption{Discourse structure graph}
\label{fig:rst}
\end{subfigure}
\caption{Graph-based representations of diagram structure in AI2D-RST. Note that AI2D-RST does not model arrowheads as diagram elements, but retrieves this information as necessary from the original AI2D annotation.}
\label{fig:ai2d_rst}
\end{figure*}

In order to pull apart how different expressive resources are combined into discourse structures in diagrams, AI2D-RST represents each diagram using three distinct graphs that correspond to three distinct but complementary layers of annotation: grouping, connectivity and discourse structure. Figure \ref{fig:ai2d_rst} exemplifies the graphs for all three annotation layers for the diagram introduced in Figure \ref{fig:orig_diag}. Figure \ref{fig:lay_seg2} shows the AI2D layout segmentation, which provides the inventory of diagram elements for AI2D-RST. The grouping graph in Figure \ref{fig:gg} organises diagram elements that are likely to be perceived as belonging together into visual perceptual groups, which are loosely based on Gestalt principles of visual perception \citep{ware2012}. The resulting organisation is represented using a hierarchical tree graph. Grouping nodes with the prefix `G' are added to the graph as parents to nodes that are grouped together during annotation. The grouping nodes can be picked up in subsequent annotation layers to refer to a group of diagram elements and thereby serve as a foundation for describing both connectivity and discourse structure layers \citep[7--8]{hiippalaetal2019-ai2d}.

The connectivity layer is represented using a cyclic graph whose edges represent visually \emph{explicit} connections that are signalled using arrows and lines and which hold between diagram elements and their groups defined in the AI2D-RST grouping layer \citep[9]{hiippalaetal2019-ai2d}. Because the rock cycle diagram in Figure \ref{fig:orig_diag} features several arrows without explicit sources and targets (see e.g. arrows A0, A1 and A5 in Figure \ref{fig:lay_seg}), the connectivity graph in Figure \ref{fig:cg} does not reflect the cyclic nature of the phenomenon represented in the diagram. This may be traced back to the insufficient segmentation of the cross-section blob B0, which was already identified as problematic when applying the original schema proposed in AI2D to the same diagram: the arrows connect sub-regions of the cross-section (see e.g. A0 and A1), but these sub-regions are not available in the inventory of diagram elements that are used to populate the connectivity graph. In the case of both AI2D and AI2D-RST, the problem originates in the stratum of expressive resources and their segmentation into analytical units. Some visual expressive resources may require more fine-grained decomposition than others, but this information becomes available only when these resources are considered as a part of the discourse structure they participate in.

This brings us to the final layer for discourse structure, which uses Rhetorical Structure Theory to describe semantic relations between diagram elements as suggested above. The discourse relations defined in RST are intended to capture the communicative intentions of the designer, as judged by an analyst, and are added to the discourse structure graph as nodes prefixed with the letter `R' as shown in Figure \ref{fig:rst}; the edges of the graph describe which role an element takes in the discourse relation, namely nucleus (`n') or satellite (`s'). The notion of \emph{nuclearity} is a key criterion in definitions of semantic relations in RST and distinguishes between units of a `text' that are considered central to the argument unfolding and units that play only a supporting role. Following the original RST definitions, AI2D-RST represents the discourse structure layer using a tree graph \citep[10--13]{hiippalaetal2019-ai2d}.

In the discourse graph shown in Figure \ref{fig:rst}, the specific rhetorical relations include \textsc{identification} (R1--R6), \textsc{cyclic sequence} (R7) and \textsc{background} (R8). Since AI2D-RST relies on the inventory of diagram elements provided by the original layout segmentation in AI2D -- a decision motivated by the need to avoid annotating everything from scratch and to maintain compatibility with AI2D -- the description of discourse semantics in AI2D-RST requires making compromises that will become evident below. Here the annotator had concluded that most text instances serve to identify what the arrows stand for, namely stages of the rock cycle. The image showing the cross-section (B0), in turn, is placed in a \textsc{background} relation to the \textsc{cyclic sequence} relation. The definition of a \textsc{background} relation \citep{mannthompson1988} states that the satellite (B0) increases the ability to understand the nucleus (R7), which is the top-level relation assigned to the diagram's representation of the entire cycle.

The analysis in Figure \ref{fig:rst} remains, however, a rather crude description of the discourse structure of the diagram in Figure \ref{fig:lay_seg}, because the cross-section actually provides far more information than what is captured by the AI2D layout segmentation, which assigns the entire cross-section into the blob B0. Just as the preceding discussion of the connectivity layer showed, insufficient analytical granularity results in an incomplete inventory of diagram elements. The information contained in these sub-regions is crucial for understanding what the diagram is attempting to communicate \emph{but we cannot know that such a decomposition into sub-regions is necessary without considering the discourse organisation of the entire diagram first}. This demonstrates why the decomposition of diagrams should be pursued in a top-down direction and be guided by the discourse structure \citep{batemanwildfeuer2014b}. This means that, even when supported by an annotation framework with sufficient local and global reach in terms of discourse structure, resulting descriptions are unlikely to be adequate unless the inventory of analytical units has been appropriately populated.


\subsection{Layout across diagram categories}
\label{sec:layout}


With two sizeable corpora with rich annotations at hand, we now turn to examine how decisions related to the metadata scheme in the AI2D dataset are propagated to the AI2D-RST corpus. As pointed out above, the two corpora differ in terms of the metadata schemes used to describe the diagrams, and this difference may be traced back to the research interests of artificial intelligence and multimodality research, and the assumptions these fields make about the sensory and semiotic nature of multimodality as a phenomenon (see Section \ref{sec:mmdh}).

We begin by focusing on \emph{layout}, a key expressive resource in the diagrammatic mode, whose importance has been underlined in both artificial intelligence and multimodality research. Research on graphic and information design has long acknowledged that document layout supports access to discourse structure, which raises the question whether layout plays a similar role in diagrams, and if so, to what extent \citep{waller2012}. \citet[149]{andrerist1995} observe that ``layout has to be considered as an important carrier of meaning'' in diagrams, because it generates hypotheses about their discourse structure \citep[see also][]{watanabenagao1998}. For the same reason, layout should be considered a prime target for distant viewing, as the information on expressive resources and their placement in the layout is readily available in the AI2D layout segmentation. In short, layout information may provide valuable clues about the communicative goals of the diagrams.

\begin{figure}[h!]
\centering
\includegraphics[width=1\textwidth]{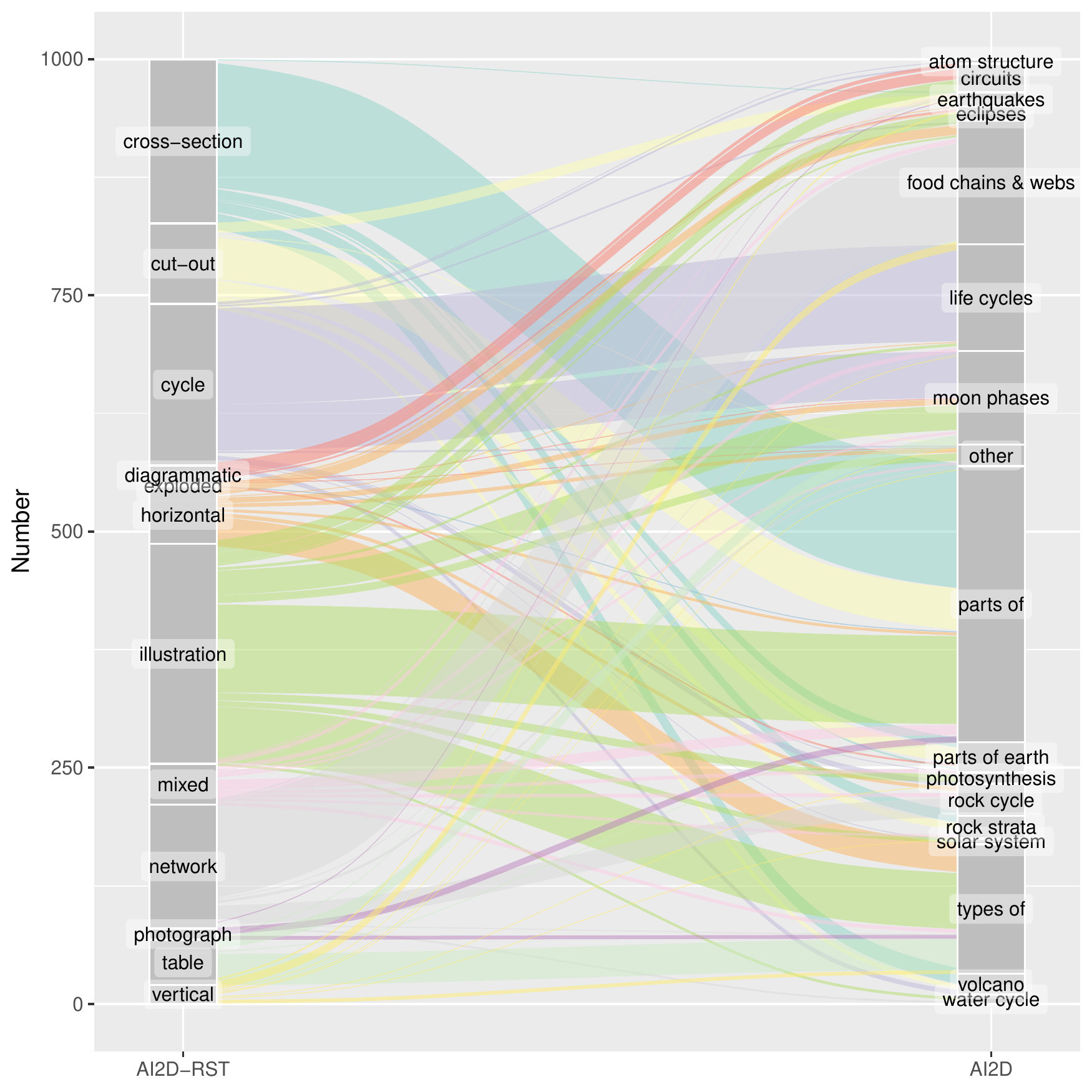}
\caption{An alluvial plot mapping the structural categories in AI2D-RST to the semantic categories in AI2D. Y-axis shows the number of diagrams in each category. Diagrams in AI2D-RST that combine multiple categories are labelled as `mixed'. Created using \emph{ggalluvial} 0.12.3 for R 4.0.2.}
\label{fig:catmap}
\end{figure}

Although both corpora share the same layout segmentation, AI2D and AI2D-RST differ in terms of how they categorise the diagrams. AI2D assigns diagrams into 17 semantic categories that mainly correspond to the subject matter of the diagram, as exemplified by \emph{volcano} and \emph{rock strata}, whereas AI2D-RST defines 11 structural categories that represent abstract diagram types, such as \emph{network}, \emph{cycle} and \emph{cross-section}. These structural categories attempt to capture patterned ways of combining expressive resources made available by the diagrammatic mode that apply independently of particular subject matters. In multimodality research, such patterns are often attributed to genre, a higher-level phenomenon that generates expectations towards the content and structure of multimodal discourse by providing structural cues that evoke previous encounters with similar communicative situations and artefacts \citep{lemke2005, bateman2008, hiippala2015a}. Figure \ref{fig:catmap} shows how the 1000 diagrams in AI2D-RST are mapped to structural categories in AI2D-RST and semantic categories in AI2D.

For the purposes of distant viewing, the semantic and structural categories in AI2D and AI2D-RST can be treated as different metadata schemes that can impose structure on the observations made for expressive resources and their placement in the layout space. To explore differences between these schemas, we retrieve the diagrams from each semantic and structural category and calculate the centroid of each blob, line and text element. Because the diagrams are of different size and orientation, we normalise the horizontal and vertical coordinates for each centroid by dividing the coordinates by the width and height of the original diagram image. We then use multivariate kernel density estimation to estimate probabilities for the position of diagram element centroids along the horizontal and vertical axes that demarcate the layout space.

\begin{figure*}[h!]
\centering
\begin{subfigure}[t]{0.48\textwidth}
\includegraphics[width=1\linewidth]{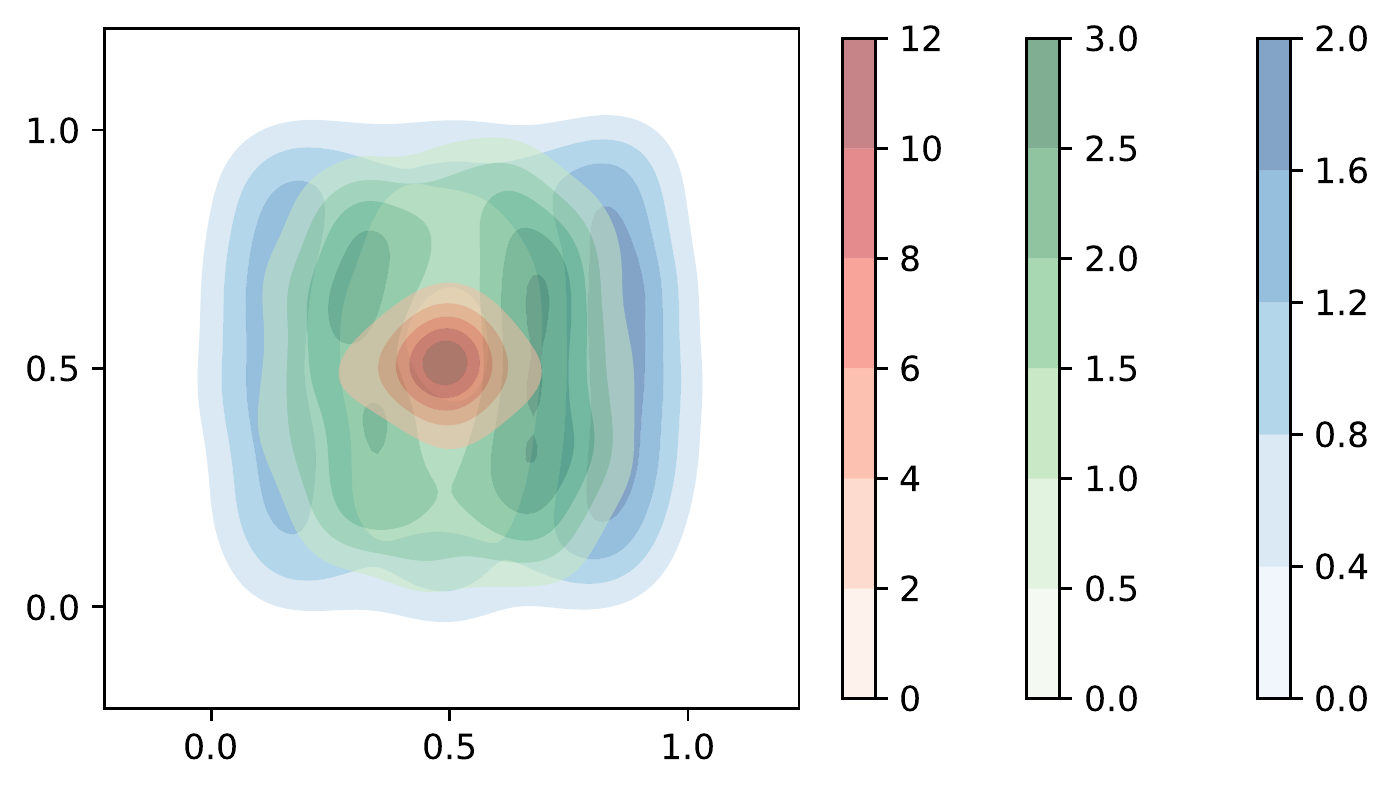}
\caption{Parts of}
\label{fig:kde_pa}
\end{subfigure}
~ 
\begin{subfigure}[t]{0.47\textwidth}
\includegraphics[width=1\linewidth]{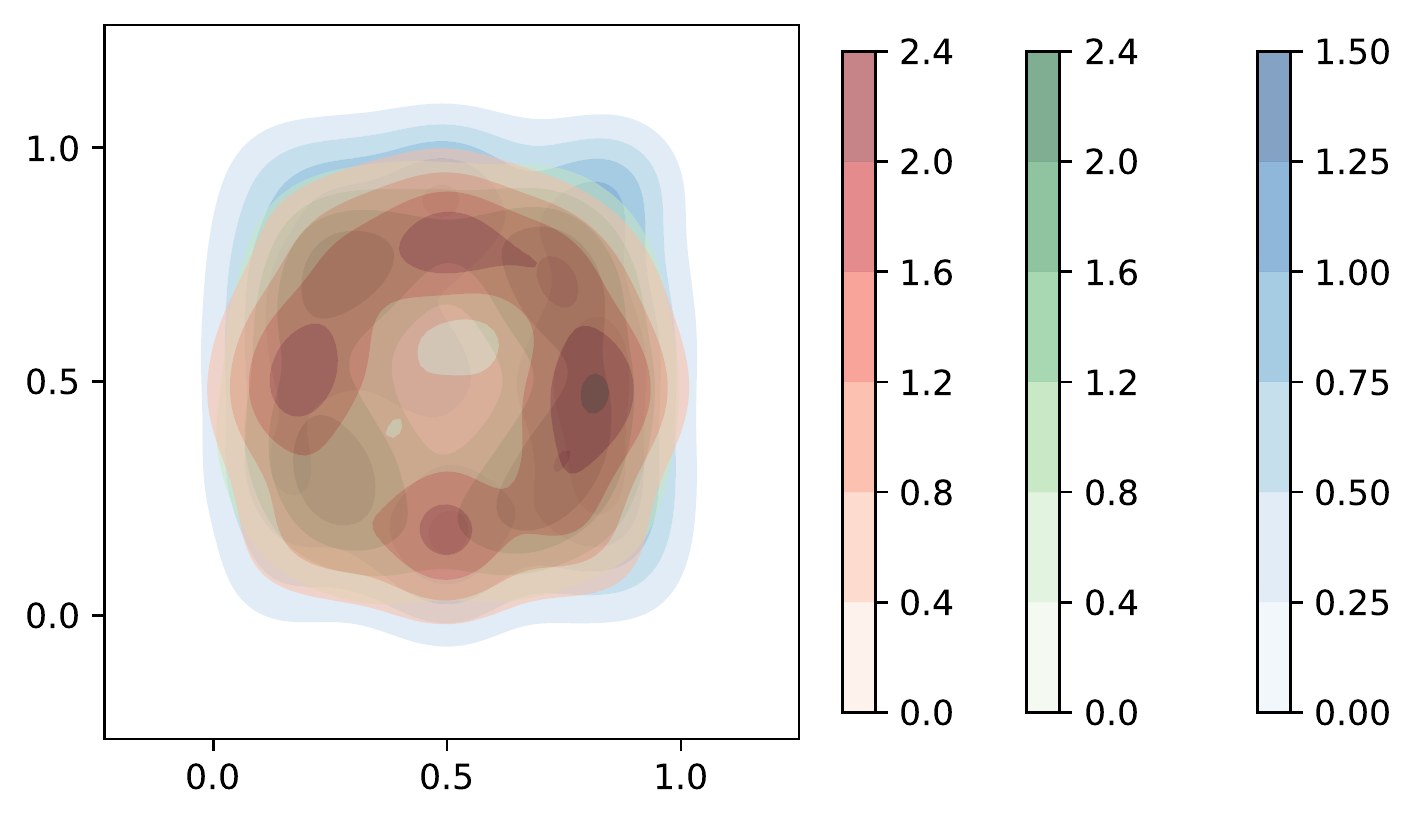}
\caption{Life cycles}
\label{fig:kde_lc}
\end{subfigure} \\[12pt]  
\begin{subfigure}[t]{0.48\textwidth}
\includegraphics[width=1\linewidth]{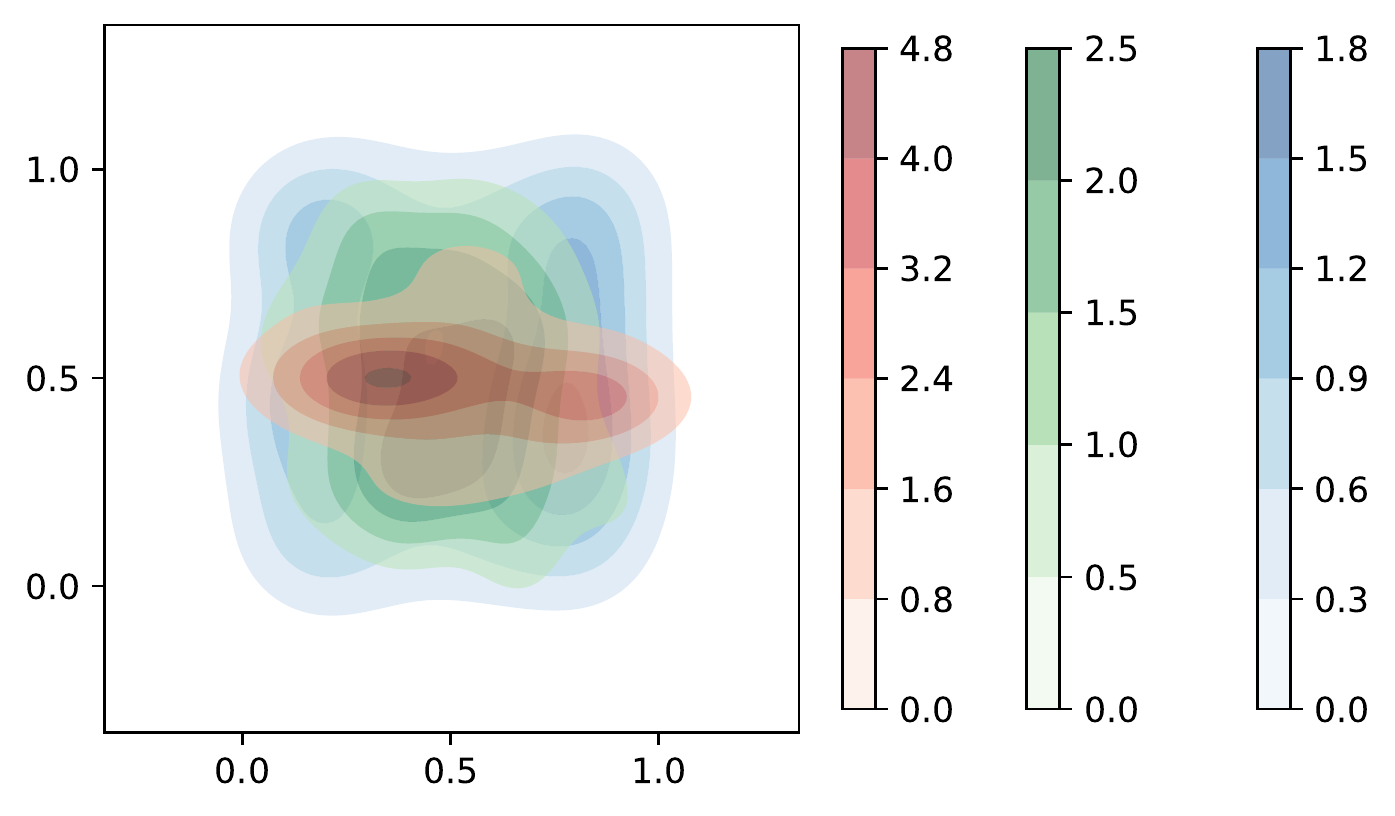}
\caption{Rock strata}
\label{fig:kde_rs}
\end{subfigure}
~ 
\begin{subfigure}[t]{0.48\textwidth}
\includegraphics[width=1\linewidth]{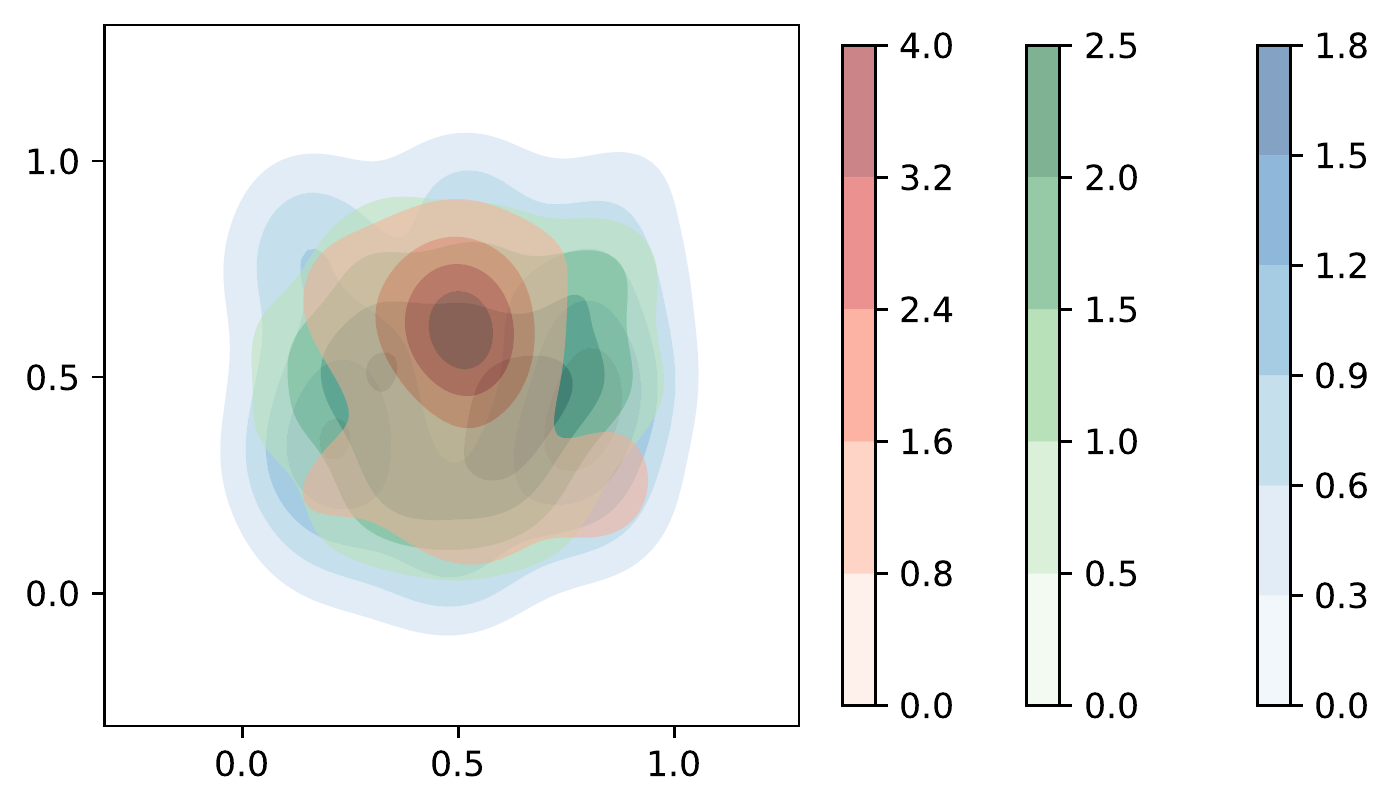}
\caption{Volcano}
\label{fig:kde_vo}
\end{subfigure}
\caption{Kernel density estimations for the centroids of text (blue), arrow/line (green) and blob (red) elements for four semantic categories in the AI2D dataset. The values for the probability density function for each diagram element type are given by the coloured bars on the right. Created using \emph{matplotlib} 3.3.0 \citep{hunter2007} and \emph{seaborn} 0.10.1 for Python 3.8.5.}
\label{fig:ai2d_kde}
\end{figure*}

Figure \ref{fig:ai2d_kde} shows kernel density estimations (KDE) for four semantic categories in the AI2D dataset: \emph{parts of}, \emph{life cycles}, \emph{rock strata} and \emph{volcano}. The KDE plot for the category \emph{parts of} in Figure \ref{fig:kde_pa} reveals that the centroids of blob elements occur with high probability in the middle of the diagram, whereas the centroids of line and text elements are distributed evenly on the sides. In this pattern, the different elements occupy distinct areas of the layout, which suggests that the objects under description are positioned in the middle and their parts are picked out using lines and written labels positioned along the outer edges. Interestingly, \emph{parts of} is one of the few categories in AI2D that is not strongly aligned with a specific subject matter, but represents a generic category that includes diagrams concerned with various topics. In the category of \emph{life cycles}, shown in Figure \ref{fig:kde_lc}, the centroids for blobs occur with higher probability in areas positioned on top, bottom, left and right of the diagram. These areas are likely to stand for particular stages of the life cycle. In contrast to the category \emph{parts of} in Figure \ref{fig:kde_pa}, blobs, lines and text do not occupy distinct areas in the layout of \emph{life cycles}. Conversely, the patterns for \emph{rock strata} and \emph{volcano} in Figures \ref{fig:kde_rs} and \ref{fig:kde_vo} respectively bear some similarities to the pattern for \emph{parts of}, but seem more variable in their positioning of diagram elements, as reflected by the spatial pattern and lower values for the probability density estimates.

\begin{figure*}[t]
\centering
\begin{subfigure}[t]{0.48\textwidth}
\includegraphics[width=1\linewidth]{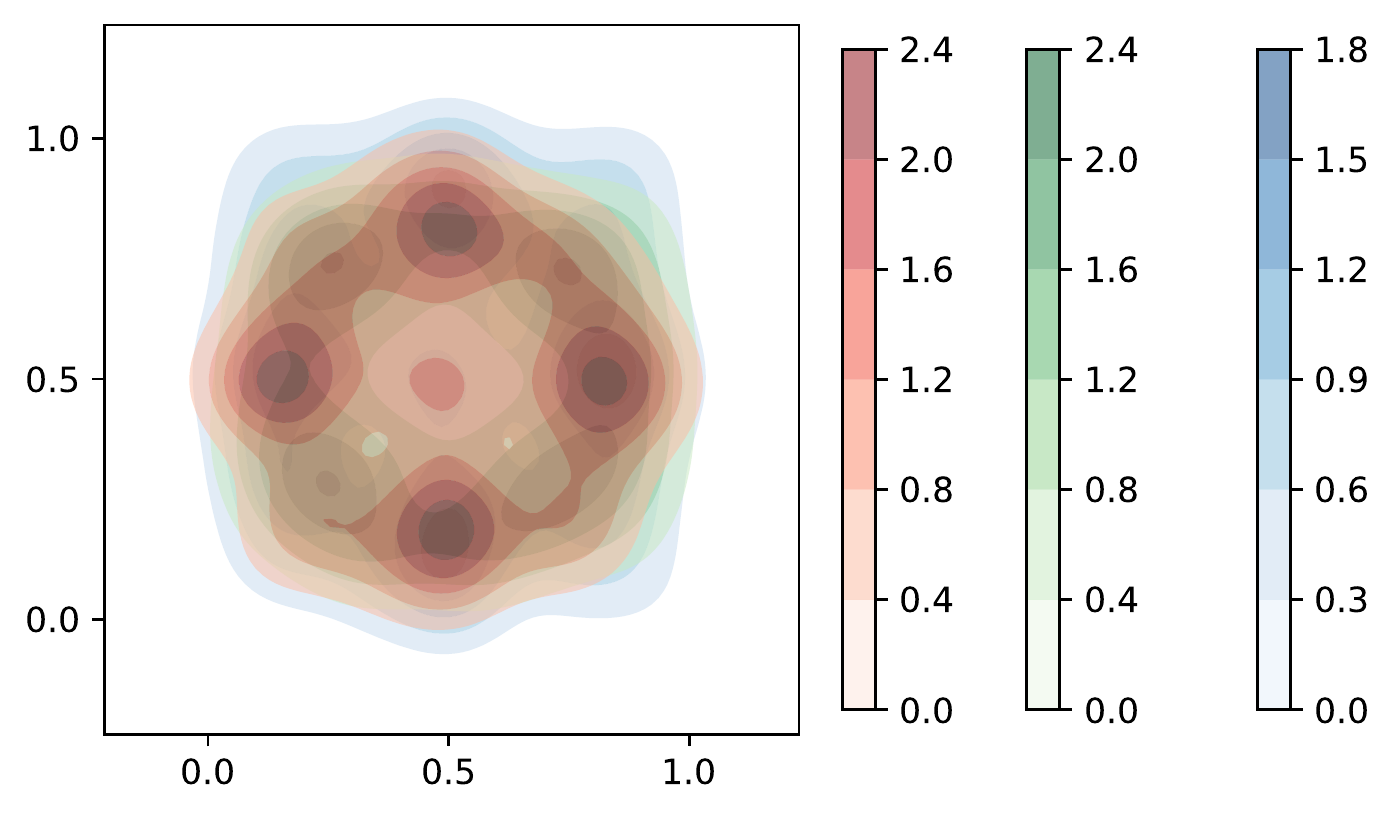}
\caption{Cycle}
\label{fig:kde_cy}
\end{subfigure}
~ 
\begin{subfigure}[t]{0.46\textwidth}
\includegraphics[width=1\linewidth]{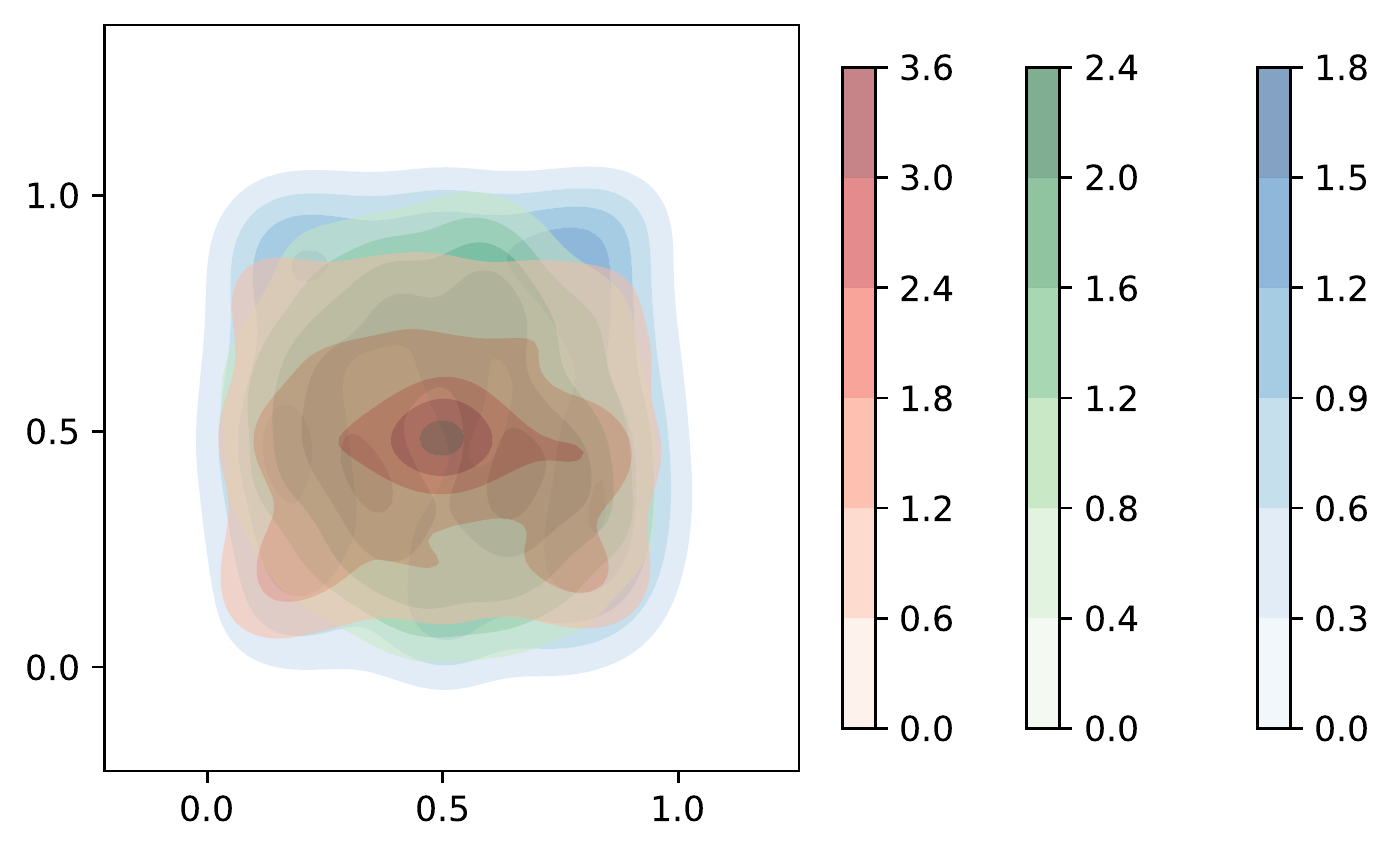}
\caption{Illustration}
\label{fig:kde_il}
\end{subfigure} \\[12pt]  
\begin{subfigure}[t]{0.48\textwidth}
\includegraphics[width=1\linewidth]{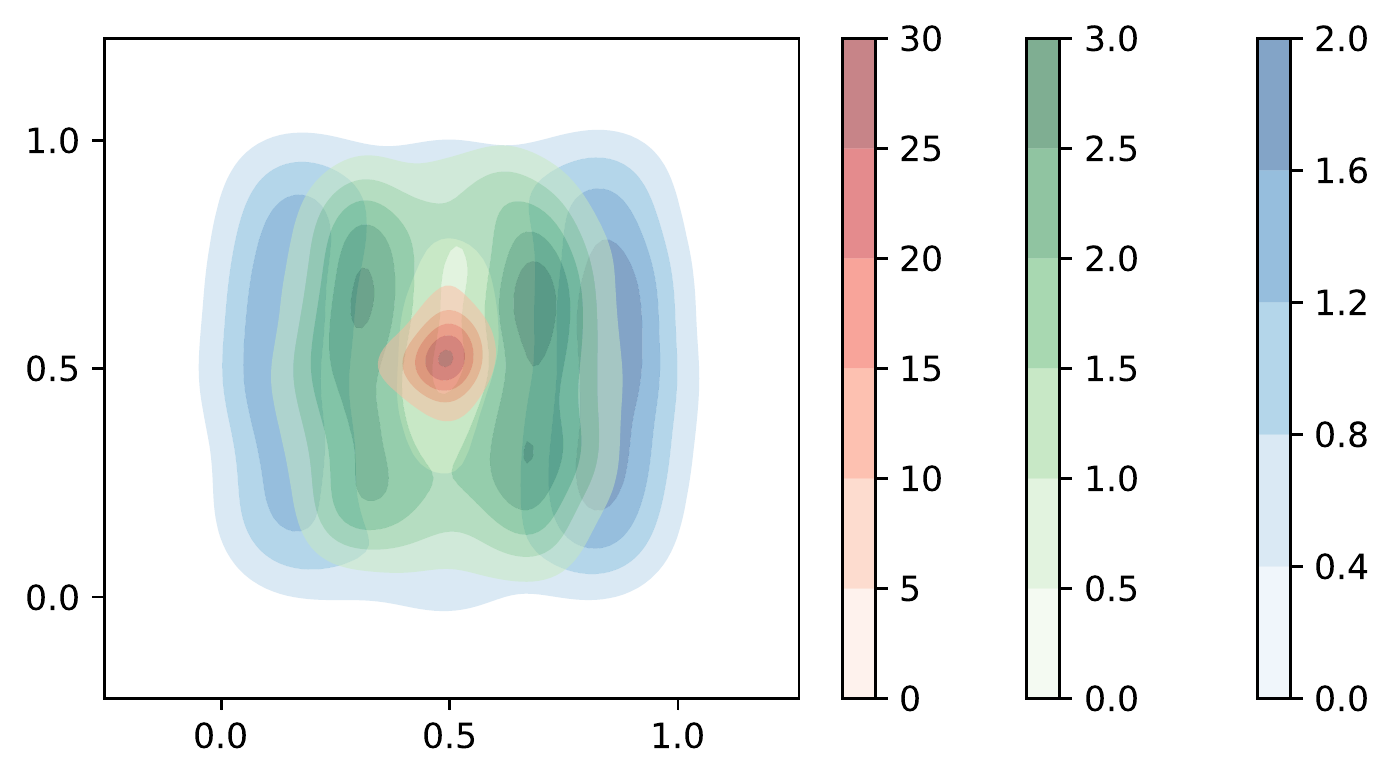}
\caption{Cross-section}
\label{fig:kde_cs}
\end{subfigure}
~ 
\begin{subfigure}[t]{0.47\textwidth}
\includegraphics[width=1\linewidth]{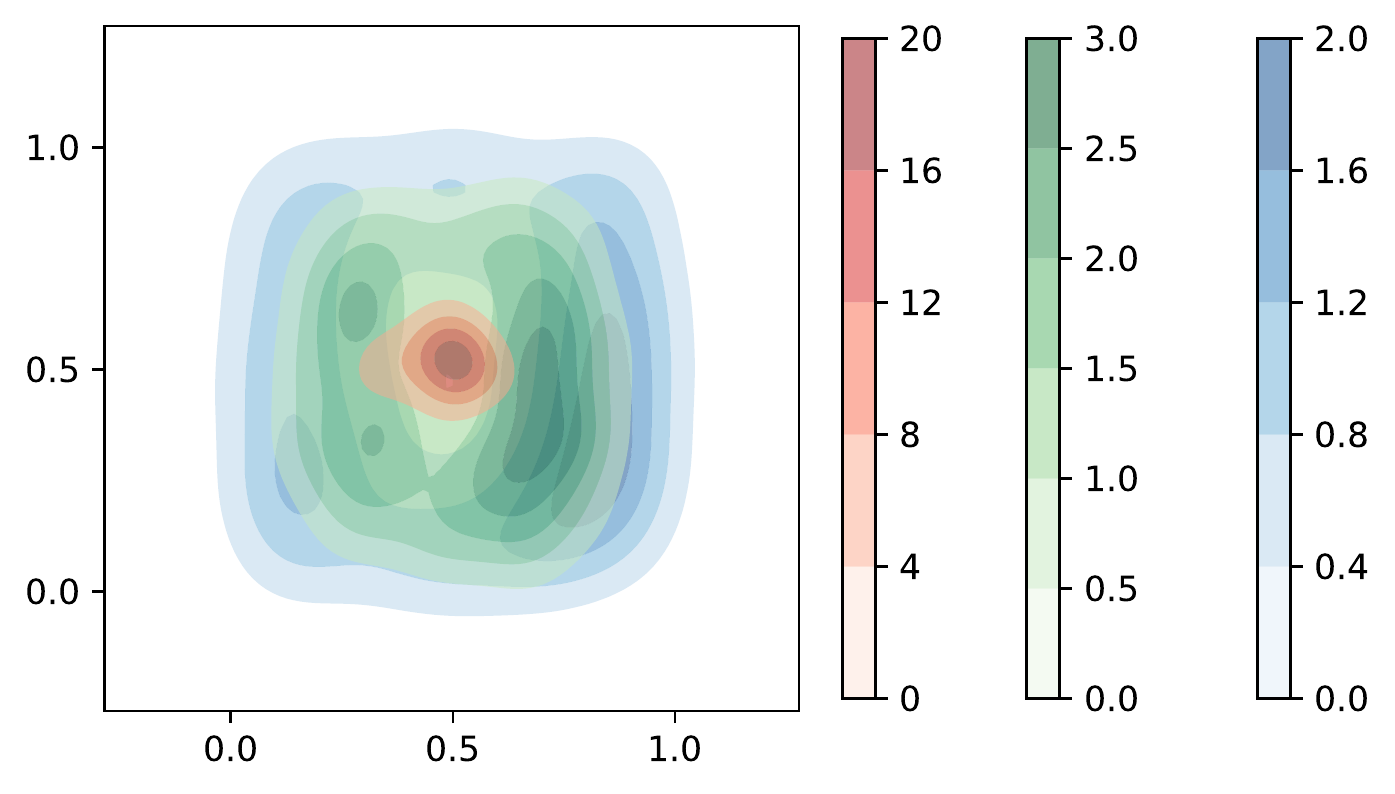}
\caption{Cut-out}
\label{fig:kde_co}
\end{subfigure} \\
\caption{Kernel density estimations for the centroids of text (blue), arrow/line (green) and blob (red) elements for four structural categories in the AI2D-RST corpus. The values for the probability density function for each diagram element type are given by the coloured bars on the right. Created using \emph{matplotlib} 3.3.0 \citep{hunter2007} and \emph{seaborn} 0.10.1 for Python 3.8.5.}
\label{fig:ai2d_rst_kde}
\end{figure*}

Figure \ref{fig:ai2d_rst_kde} shows patterns for four of the structural categories defined in AI2D-RST. The pattern for \emph{cycle} (Figure \ref{fig:kde_cy}) suggests that structural categories in AI2D-RST can bring out regularities in how diagrams use the layout space regardless of their subject matter. As the alluvial plot in Figure \ref{fig:catmap} showed, the AI2D-RST category \emph{cycle} includes diagrams from the AI2D categories \emph{life cycles}, \emph{moon phases}, \emph{rock cycle} and \emph{water cycle}. What is particularly striking about the layout pattern for \emph{cycle} is that four major positions may be identified for blobs -- namely top, bottom, left and right -- whereas minor positions are visible between them. It may be suggested that this pattern stands out precisely because the \emph{cycle} category aggregates layout information from all diagrams that represent phenomena as cyclic, regardless of their subject matter.

The remaining examples in Figure \ref{fig:ai2d_rst_kde} show layout patterns for three structural categories in AI2D-RST concerned with how diagrams represent their depicted objects \citep[8--9]{hiippalaetal2019-ai2d}. These include \emph{illustration} in Figure \ref{fig:kde_il}, which covers all graphic representations of objects at various levels of visual detail from monochrome to colour drawings. \emph{Illustration} is distinguished from \emph{cross-section} in Figure \ref{fig:kde_cs} and \emph{cut-out} in Figure \ref{fig:kde_co} based on whether the internal structure of an object is depicted by cutting the object in half (\emph{cross-section}) or by removing a part of the object to expose its structure (\emph{cut-out}). As the layout patterns show, \emph{illustrations} are more flexible in their positioning of blobs than \emph{cross-sections} and \emph{cut-outs}, as \emph{illustrations} occasionally depict multiple objects in a single diagram which causes the blob centroids to spread out. The layout patterns for \emph{cross-sections} and \emph{cut-outs}, in turn, are so similar that they cannot be distinguished from one another without information on the structural category in question.

This is precisely where AI2D-RST begins to expose limitations set by the metadata schema originally used for describing diagram elements in the AI2D dataset. Because AI2D classifies all graphic expressive resources as `blobs', we cannot tease out whether \emph{illustrations}, \emph{cycles} and \emph{cut-outs} prefer different expressive resources for depiction, such as coloured hand-drawn illustrations or monochrome line drawings. As we argued above in Section \ref{sec:ai2d}, defining the expressive resources ahead of analysis limits the ability to describe what the diagrammatic mode does with a material substrate in terms of expressive resources. Identifying the expressive resources deployed is crucial for describing the operation of the diagrammatic mode, because diagrams use expressive resources to adjust their level of abstraction \citep{dimopoulosetal2003}. In other words, one needs to identify the expressive resources to recognise the kind of diagram in question.

\subsection{Unpacking the graphic expressive resources}
\label{sec:expr}

\begin{figure}[h!]
\centering
\includegraphics[width=1\textwidth]{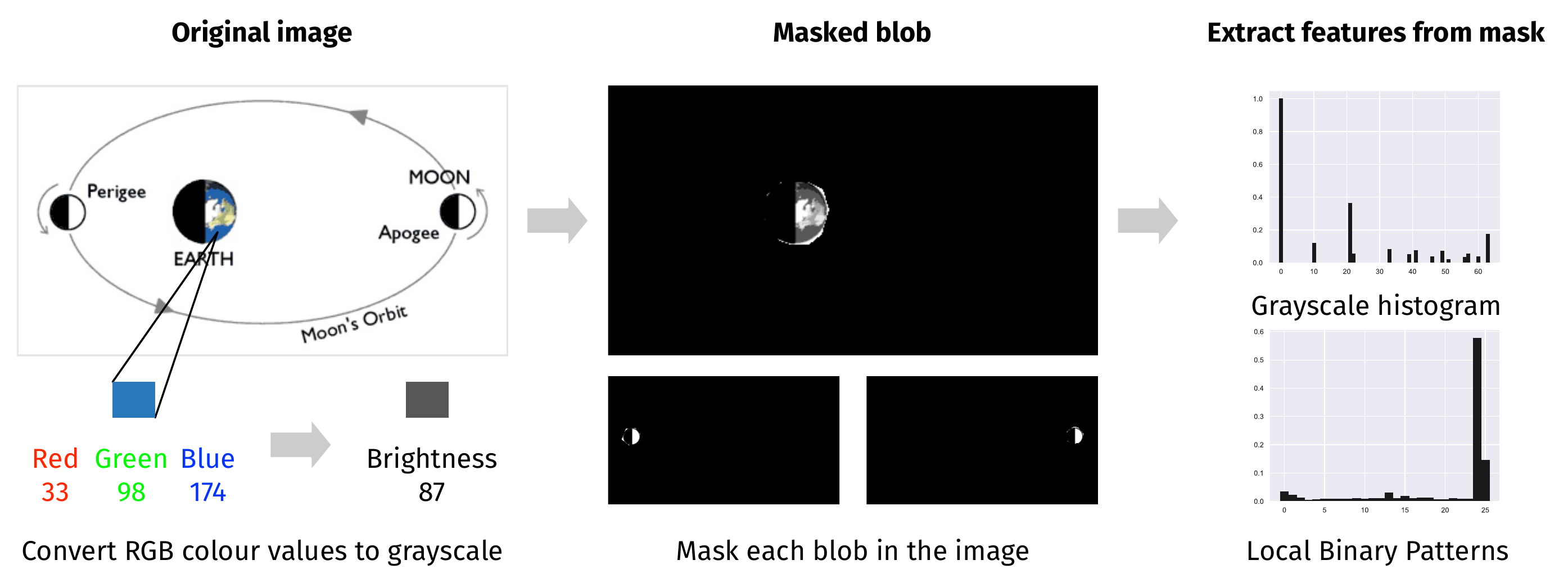}
\caption{Extracting features for brightness and texture from `blobs' in the AI2D dataset.}
\label{fig:method}
\end{figure}

For these reasons, we now pursue a more fine-grained description of the expressive resources collectively labelled as `blobs' in the AI2D dataset. To explore which expressive resources the diagrams use for depiction, we extract all diagram elements classified as blobs from the AI2D dataset ($N=20937$) and apply the method presented in Figure \ref{fig:method}. To characterise the visual appearance of each blob, we first convert each blob to grayscale. Each pixel in a grayscale image is represented by a value between 0 and 255, which encodes its brightness: 0 stands for black, whereas 255 stands for white. We then describe the brightness of the entire blob by calculating a grayscale histogram with 64 bins, in which each bin covers a given range of pixel values. To exemplify, in this case, the first of the 64 bins covers values from zero to three. If the value of a pixel falls within this range, the value for the first bin increases by one. Placing all the pixels into the 64 bins provides a 64-dimensional vector that describes blob brightness.

We then extract Local Binary Patterns \citep[LBP: ][]{ojalaetal1996} to represent the texture of each blob using the implementation for uniform LBP in the \emph{scikit-image} library for Python \citep{vanderwaltetal2014}. LBP examines the neighbourhood of each pixel within a pre-specified window and encodes information about the pixel neighbourhood using binary values: if the value of a neighbouring pixel is lower than the value of the current pixel, the neighbour receives a value of zero. Conversely, if the value is larger, the neighbouring pixel receives a value of one. LBP then identifies binary patterns across the entire blob and aggregates this information into a histogram. We use LBP to examine 8 neighbouring pixels within a radius of 3 pixels from the centre pixel, which provides a 26-dimensional vector for representing blob texture. Finally, we concatenate the vectors for brightness and texture into a 90-dimensional feature vector for each blob in the AI2D dataset.  

\begin{figure}[p]
\includegraphics[width=1\linewidth]{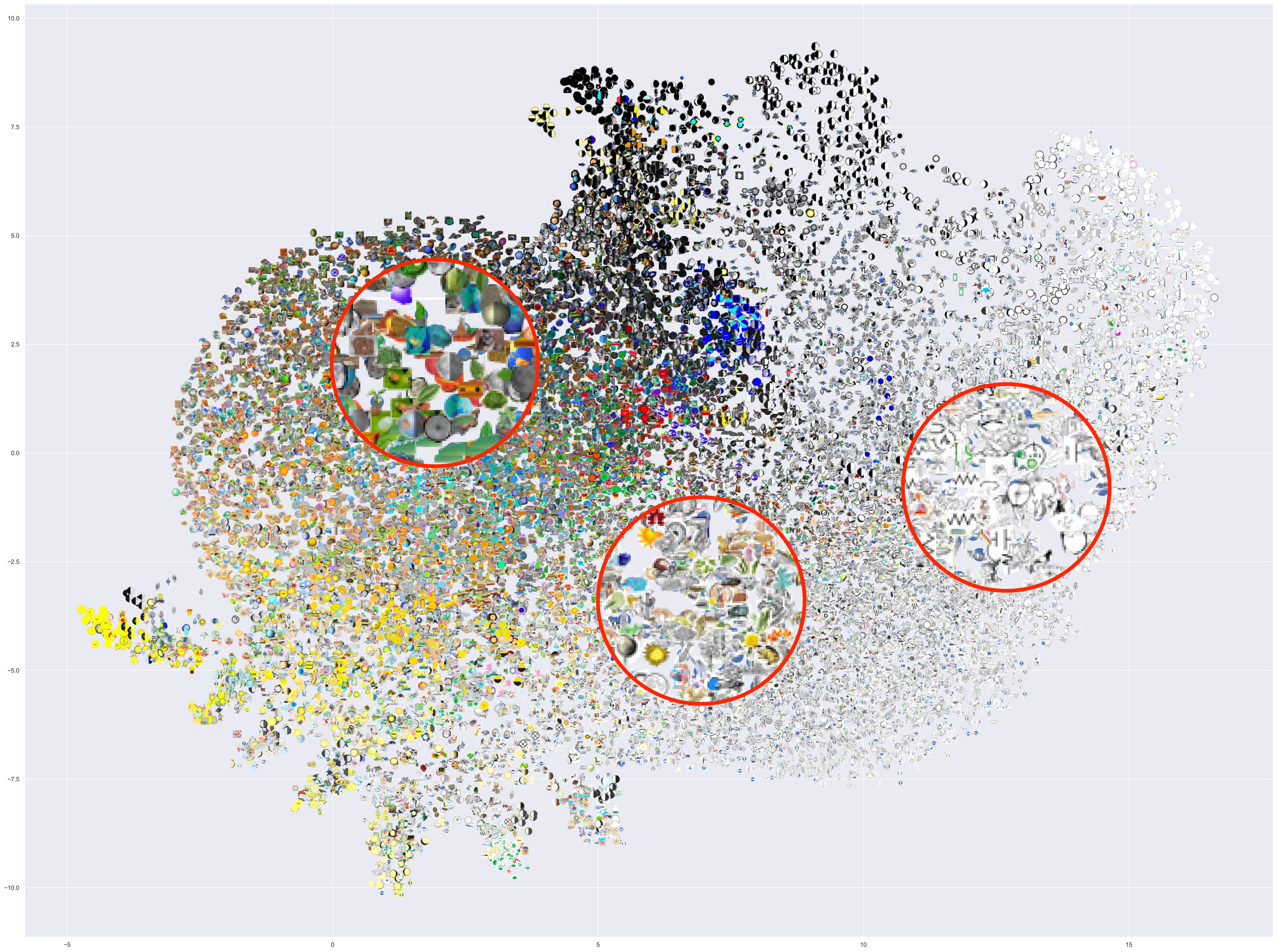}
\caption{Mapping the 90-dimensional feature space for gray histogram and local binary patterns to two dimensions for plotting using UMAP. Each blob is represented by its thumbnail. The three loupes demarcated using red colour provide zoom into different regions of the plot. Created using the \emph{matplotlib} 3.3.0 \citep{hunter2007} and \emph{seaborn} 0.11.0 libraries for Python 3.8.5.}
\label{fig:umap_thumbnail}
\end{figure}

We then project the 90-dimensional feature vectors into a two-dimensional space for visual exploration. To do this, we use the Uniform Manifold Approximation and Projection (UMAP) algorithm to learn a mapping between the 90- and two-dimensional feature spaces \citep{mcinnes2018-software}. UMAP is controlled by two parameters: nearest neighbours and minimum distance. We set nearest neighbours to $200$, which seeks to emphasise global over local structure when determining neighbours in the high-dimensional space. For minimum distance, we set the value to $0.99$ to allow loose clustering of points in the low-dimensional space, which attempts to preserve the broad topological structure of the high-dimensional space. Figure \ref{fig:umap_thumbnail} shows the resulting visualisation, which plots the two-dimensional UMAP features against each other, using thumbnail images of the blobs to indicate their position in the two-dimensional space.

\begin{figure*}[p]
\centering
\begin{subfigure}[t]{0.31\textwidth}
\includegraphics[width=1\linewidth]{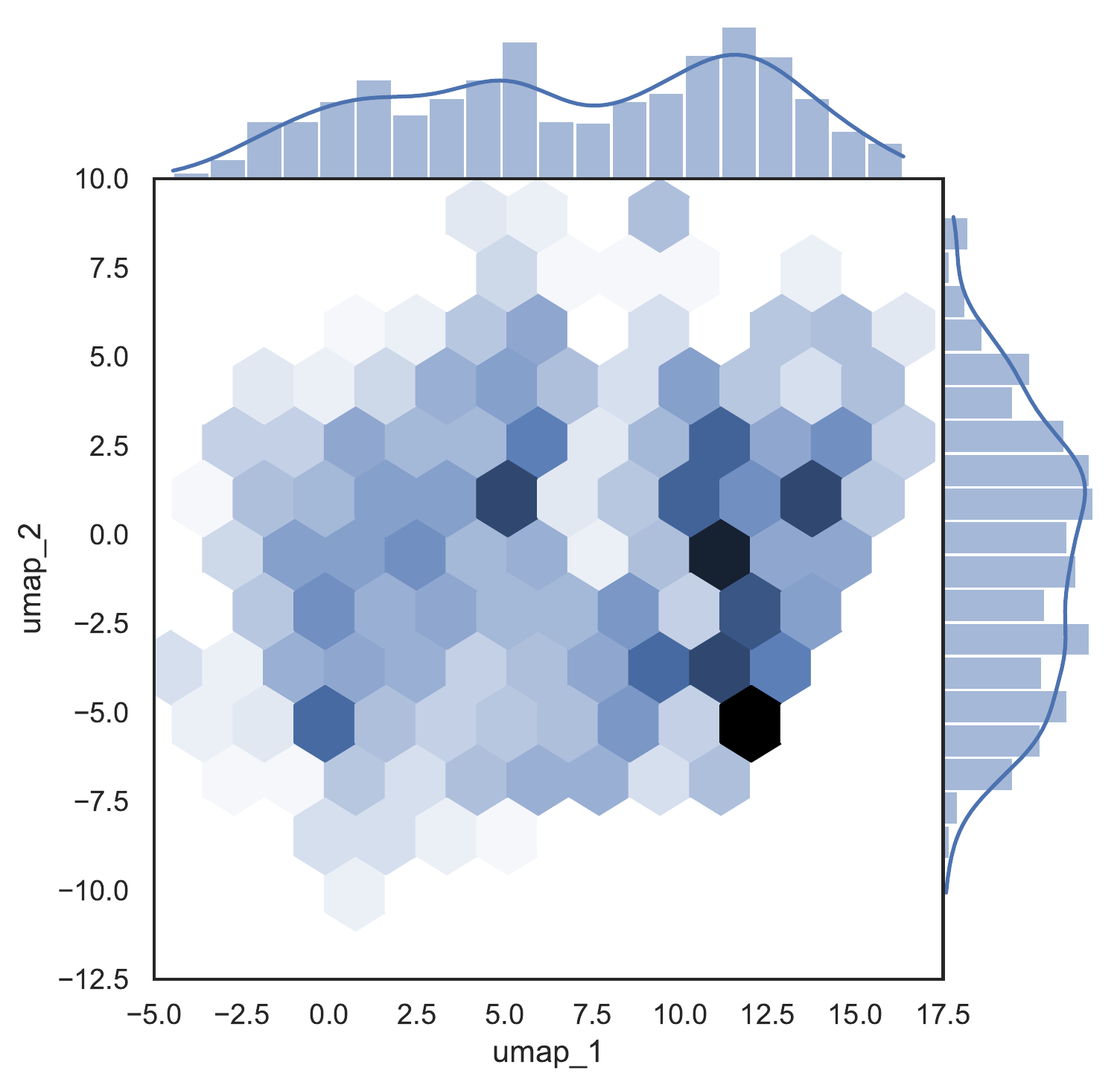}
\caption{Illustration}
\label{fig:hex_il}
\end{subfigure}
~ 
\begin{subfigure}[t]{0.31\textwidth}
\includegraphics[width=1\linewidth]{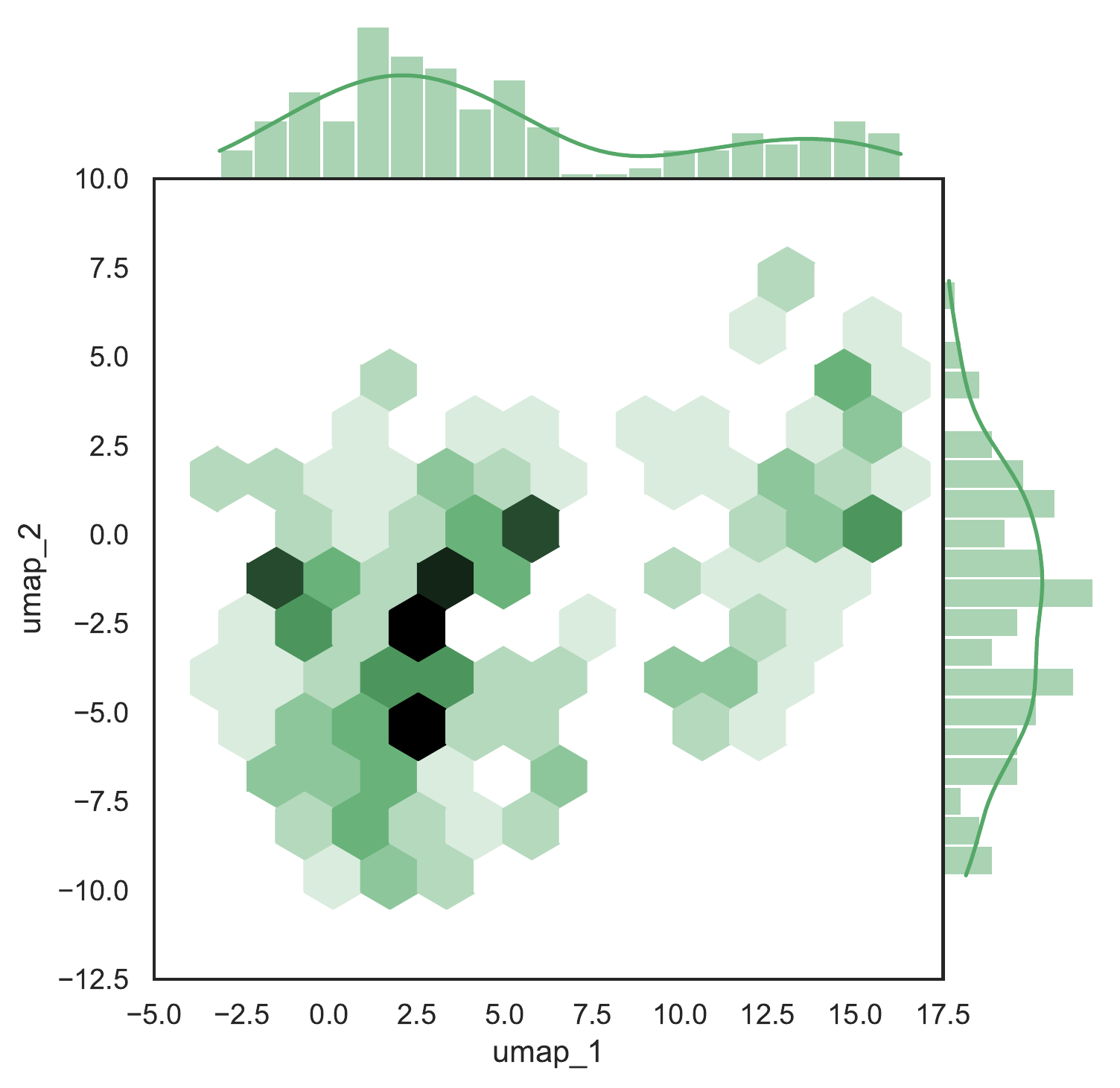}
\caption{Cross-section}
\label{fig:hex_cs}
\end{subfigure}
~ 
\begin{subfigure}[t]{0.31\textwidth}
\includegraphics[width=1\linewidth]{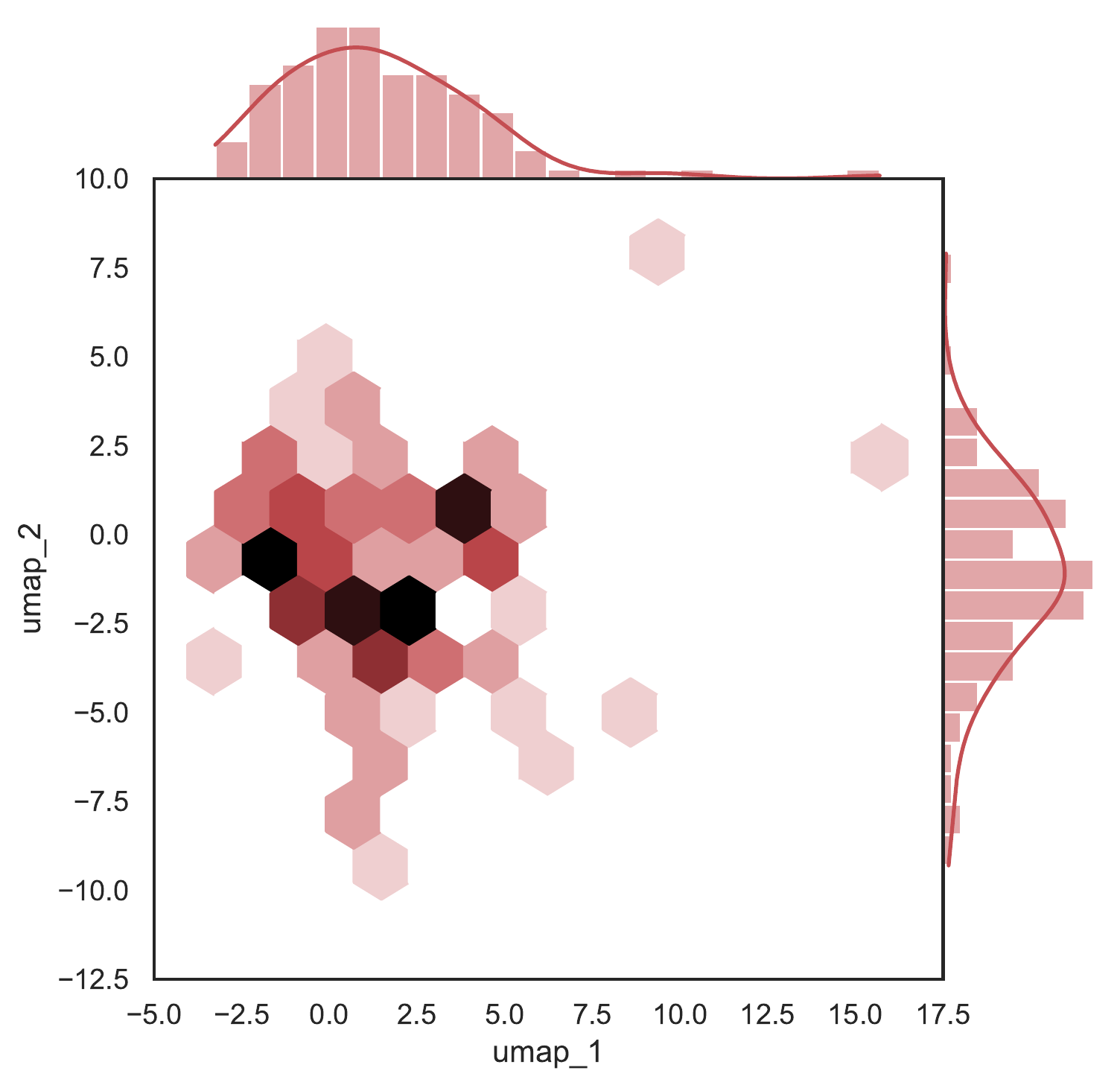}
\caption{Cut-out}
\label{fig:hex_co}
\end{subfigure}
\caption{The distribution of blob centroids across three structural categories in AI2D-RST. The marginal plots show histograms with 20 bins and a kernel density estimation for each UMAP dimension. Created using the \emph{matplotlib} 3.3.0 \citep{hunter2007} and \emph{seaborn} 0.11.0 libraries for Python 3.8.5.}
\label{fig:expr_hex}
\end{figure*}

Figure \ref{fig:umap_thumbnail} reveals that the diagram element of `blobs' covers a wide range of expressive resources. A dimension of variation may be identified along the horizontal axis, which encodes colour information: coloured drawings gradually turn into monochrome line drawings when moving from left to right. Another dimension of variation exists along the vertical axis: blobs with solid colours and texture are positioned along the outer edges, while the level of visual detail increases towards the middle. Although UMAP preserves enough information about colour and texture to capture colour differentiation (i.e., number of colours) and modulation (i.e., the range of colour shades), Figure \ref{fig:umap_thumbnail} is not sufficient for distinguishing between modes of depiction, e.g., whether the blob represents an illustration, cross-section or cut-out and \emph{which} expressive resource is mobilised for depiction. 

To better understand how the expressive resources visualised in Figure \ref{fig:umap_thumbnail} are associated with particular types of depiction captured by the structural categories in AI2D-RST (e.g. \emph{illustration}, \emph{cross-section} and \emph{cut-out}), Figure \ref{fig:expr_hex} shows three hex plots that indicate the distribution of blob centroids along the horizontal (colour) and vertical (visual detail) dimensions just for these three categories. In terms of colour, Figure \ref{fig:hex_il} reveals that blobs in \emph{illustrations} are distributed evenly along the horizontal dimension, but peak in the region corresponding to monochrome line drawings. For \emph{cross-sections} shown in Figure \ref{fig:hex_cs}, blobs are more likely to occur in the region for drawings with colour gradients along the horizontal axis, but some can also be found in the region for black and white line drawings. Blobs from \emph{cut-outs}, in turn, are very likely to be found in the region for coloured drawings, as shown in Figure \ref{fig:hex_co}. In terms of visual detail, blobs from all three categories peak around the region for visual detail, but especially strongly for \emph{cut-outs}.

\emph{Cut-outs} clearly prefer coloured drawings with rich visual detail that are needed to depict three-dimensional objects and their structure. This suggests that Figure \ref{fig:expr_hex} reflects functionally motivated choices related to graphic expressive resources. In other words, the way objects are being depicted influences the choice of expressive resources. Notably, \emph{cross-sections} draw on both coloured and monochrome drawings, which are both suitable for depicting the internal structure of objects from their side. \emph{Illustrations}, in turn, are more flexible in terms of their choice of graphic expressive resources. These results are generally aligned with the findings of \citet[200]{dimopoulosetal2003}, who manually annotated over 2800 primary school science diagrams for four features that affect the degree of abstraction: (1) the presence of geometrical shapes and symbols, (2) the variety of colours used, (3) their range of shades, and (4) contextualisation, that is, whether the depiction uses an illustrated or plain colour background. Given that our method produces similar insights without using annotations for these particular features suggests that the generic metadata schema indeed captures key characteristics of the diagrammatic mode in this domain, such as the selection of expressive resources, which may be linked back to particular communicative needs.

\section{Discussion}

For the analysis of diagram layouts in Section \ref{sec:layout}, the results emphasise the flexibility of the diagrammatic mode: the same structural configuration can be used to realise diagrams that deal with different topics, as long as the phenomena represented share certain common features (cf. Figure \ref{fig:catmap}). We characterised these structural configurations in terms of genre, which refers to high-level patterns of expressive resources \citep[cf.][]{bateman2008}. Our results show that these genre-based structural categories appear to correspond to high-level layout patterns, as shown in Figure \ref{fig:ai2d_rst_kde}. However, \citet[21--22]{hiippalaetal2019-ai2d} observe that genre patterns are fluid and a single diagram may incorporate high-level patterns associated with multiple genres. Therefore, the layout patterns in Figure \ref{fig:ai2d_rst_kde} should not be seen as representing fixed templates for diagram design, but reflect instead conventional ways of organising expressive resources in the 2D layout space motivated by communicative functions. The degree to which these patterns are conventionalised varies from one genre to another.

In terms of the expressive resources analysed in Section \ref{sec:expr}, our analysis revealed just how much of the variation is hidden when an under-differentiating label such as `blob', `image' or `visual' is employed. The results also underline the difference between photographic images and other means of graphic expression, which have radically different capabilities for representation. Diagrams regularly draw on graphic expressive resources that allow adjusting the representational `accuracy' to match their communicative needs \citep[cf.][]{dimopoulosetal2003, greenberg2018}. In terms of methodology, it is worth noting that extracting features from blobs using a convolutional neural network pre-trained on \emph{photographs} did not produce meaningful UMAP clusters. \citet[200]{weverssmits2019} report on similar experiences when applying pre-trained neural networks to historical newspapers, where multiple graphic modes of expression are used alongside photographs. This suggests that many computer vision techniques that work well on photographs are not directly transferrable to other graphic modes of expression. 

Generally, our results underline the importance of paying attention to modelling expressive resources when building corpora for specific semiotic modes. Without describing the expressive resources at a sufficient degree of accuracy and granularity, the analysis will not be able to capture what the semiotic mode does with its material substrate, nor to provide an inventory of discourse units needed to represent its discourse semantics appropriately. Just which expressive resources are made available by a semiotic mode must be answered through empirical research. As \citet{thomas2014} notes, interpretation -- or in this case, making decisions concerning particular expressive resources -- should be motivated by observations made in the data and delayed until unavoidable. Exploring expressive resources using the techniques proposed in Section \ref{sec:expr} may be used to support this process.

What the results also show is that the concept of a semiotic mode can be applied productively to creating corpora and training data for digital humanities and artificial intelligence research because it adds specifically significant structures over the measured data. There is much potential in using the stratum of discourse semantics as the basis for defining crowdsourcing tasks alongside common strategies adopted in computer vision research \citep[cf.][]{kovashkaetal2016}. Rather than requesting the workers to identify and place elements into pre-defined categories, the crowdsourcing tasks could be used to tease out discourse interpretations related to the data and convert them into descriptions of discourse structures. To exemplify, if a diagram features an element that consists of written text, it would be natural to ask whether the text refers to a particular entity in the diagram. This would allow injecting aspects of the inherently dynamic nature of discourse interpretations into the annotation tasks.

Finally, we argue that our analysis shows that pre-theoretical distinctions between `text' and `image' rarely hold in multimodal communication, and they are seriously under-differentiating for the digital humanities or any other field concerned with multimodal visual expression. While we fully agree with \citet{arnoldtilton2019} on the need to develop methods that enable the large-scale analysis of various media, we have demonstrated here that this effort must be supported by a solid theoretical foundation that reveals rather than hides the complexity of multimodal communication \citep{bateman2017b}. This foundation is needed for tackling issues that are traditionally of concern to the humanities, such as trajectories of change over time, whose computational analysis is still largely limited to linguistic material. 

\section{Conclusion}

In this article, we have argued that contemporary theories of multimodality can inform computational approaches to the analysis of multimodal communication in digital humanities and beyond. Our analysis of two multimodal corpora consisting of primary school science diagrams showed how multimodality theory can reveal descriptive shortcomings that lead to serious analytical blind spots. We thus proposed that any corpus-building or modelling effort should be informed by multimodality theory and proceed in a top-down direction. This direction emphasises the need to focus on communicative intentions, as reflected in discourse structure, which determines the extent to which individual semiotic modes and their combinations must be decomposed to achieve a sufficiently coherent description of multimodal discourse. Achieving such descriptions for the wide range of historical and contemporary media requires a large-scale effort supported by expertise from researchers in humanities and computer science.

\end{document}